\documentclass[10pt,twocolumn,letterpaper]{article}

\usepackage{iccv}
\usepackage{times}
\usepackage{epsfig}
\usepackage{graphicx}
\usepackage{amsmath}
\usepackage{amssymb}
\usepackage{todonotes}
\definecolor{ForestGreen}{RGB}{34,139,34}
\usepackage[pagebackref=true,breaklinks=true,letterpaper=true,colorlinks,bookmarks=false]{hyperref}

\iccvfinalcopy % *** Uncomment this line for the final submission

\def\iccvPaperID{25} % *** Enter the ICCV Paper ID here

\ificcvfinal\fi

\usepackage{microtype}
\usepackage{graphicx}
\usepackage{booktabs} % for professional tables
\usepackage{subcaption}
\usepackage{graphicx}
\usepackage{amsmath}
\usepackage{amssymb}
\usepackage{mathtools}
\usepackage{amsthm}

\usepackage{array}
\usepackage{multicol}
\usepackage{wrapfig}
\usepackage{multirow}
\usepackage{colortbl}
\usepackage{dblfloatfix}
\definecolor{gray9}{gray}{.9}
\definecolor{gray95}{gray}{.95}
\definecolor{gray8}{gray}{.8}
\definecolor{gray85}{gray}{.85}
\begin{document}

\title{DatasetEquity: Are All Samples Created Equal? \\In The Quest For Equity Within Datasets}
% It is OKAY to include author information, even for blind
% submissions: the style file will automatically remove it for you
% unless you've provided the [accepted] option to the icml2022
% package.

\author{Shubham Shrivastava\thanks{Correspondence to: \texttt{\url{shubham@TowardsAutonomy.com}}}, Xianling Zhang, Sushruth Nagesh, Armin Parchami\\
Ford Motor Company\\
{\tt\small \{sshriva5, xzhan258, snagesh1, mparcham\}@ford.com}}

\maketitle
% Remove page # from the first page of camera-ready.
\ificcvfinal\thispagestyle{empty}\fi

\begin{abstract}

Data imbalance is a well-known issue in the field of machine learning, attributable to the cost of data collection, the difficulty of labeling, and the geographical distribution of the data. In computer vision, bias in data distribution caused by image appearance remains highly unexplored. Compared to categorical distributions using class labels, image appearance reveals complex relationships between objects beyond what class labels provide. Clustering deep perceptual features extracted from raw pixels gives a richer representation of the data. This paper presents a novel method for addressing data imbalance in machine learning. The method computes sample likelihoods based on image appearance using deep perceptual embeddings and clustering. It then uses these likelihoods to weigh samples differently during training with a proposed \textbf{Generalized Focal Loss} function. This loss can be easily integrated with deep learning algorithms. Experiments validate the method's effectiveness across autonomous driving vision datasets including KITTI and nuScenes. The loss function improves state-of-the-art 3D object detection methods, achieving over 200\% AP gains on under-represented classes (Cyclist) in the KITTI dataset. The results demonstrate the method is generalizable, complements existing techniques, and is particularly beneficial for smaller datasets and rare classes. Code is available at: 
\url{https://github.com/towardsautonomy/DatasetEquity}

% Data imbalance is a well-known issue in the field of machine learning, attributable to the cost of data collection, the difficulty of labeling, and the geographical distribution of the data. In computer vision, bias in data distribution caused by image appearance remains highly unexplored. This paper presents a novel method for addressing data imbalance in the field of machine learning, To validate our method's generalization capability, across different AV vision datasets, we present not only the qualitative data clustering results, but also the quantitative improvements in downstream perception tasks.
% Specifically in the context of 3D object detection in computer vision, various SOTA algorithms of different backbone and losses are explored. The proposed method involves weighing each sample differently during training according to its likelihood of occurrence within the dataset, \LZ{TODO: Be more specific, x percentage of improvements on KITTI} which improves the performance of state-of-the-art 3D object detection methods in terms of NDS and mAP scores. 
% The effectiveness of the proposed loss function, called \textit{Generalized Focal Loss}, was tested on two autonomous driving datasets using three different camera-based 3D object detection methods. The results show that the loss function is generalizable, complementary to existing bias mitigation techniques, and particularly effective for smaller datasets and under-represented object classes. Code will be released in the camera-ready version.
% Project page with anonymous code is available at: 
% \url{https://datasetequity.github.io/}
% \url{https://github.com/towardsautonomy/DatasetEquity}

\end{abstract}

\begin{figure*}[h]
\begin{center}
% \fbox{\rule{0pt}{2in} \rule{.9\linewidth}{0pt}}
\includegraphics[width=0.9\linewidth]{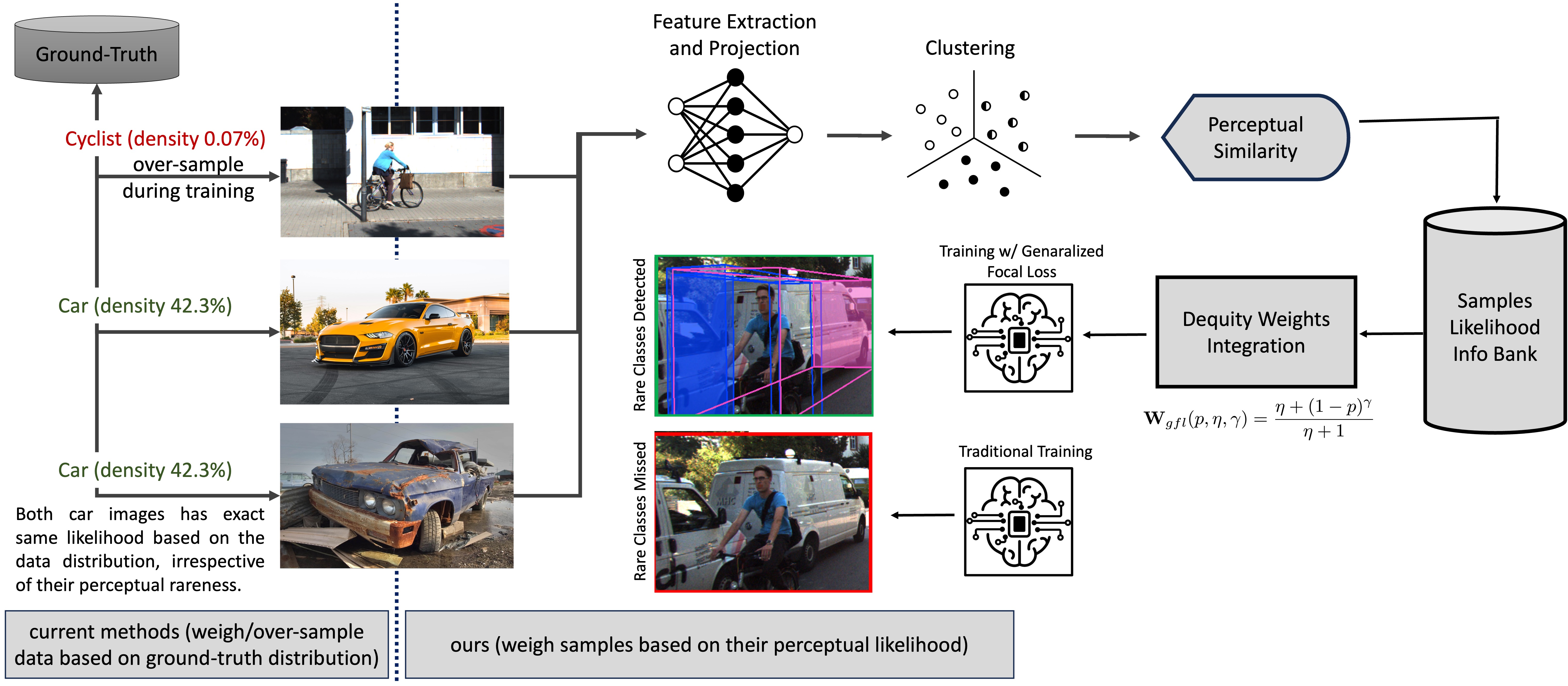}
\end{center}
    % \caption{We embark on a fascinating journey to unravel the hidden secrets within data, using the powerful way of computing sample likelihood in any dataset without requiring ground-truth information. Given these sample likelihoods, we compute \textit{Generalized Focal Loss} and train the deep learning models with this new objective function. What we find is that our loss function augments the existing methods of dealing with \textit{out-of-distribution} datasets, and improves model performance, especially on rare samples. The major advantage of our method over existing work is that the latter necessitates having class distribution available during training for under/over-sampling and/or inserting objects for augmentation, whereas ours do not have any ground-truth dependencies.}
\vspace{-0.2in}
   \caption{Left: existing techniques requiring class distribution annotations. Right: our method does not have ground truth dependencies. This work presents a method to estimate sample likelihoods by clustering semantic embeddings of raw image pixels. This enables modeling of visual relationships and data heterogeneity without reliance on categorical labels. The computed likelihoods are used to reweight training losses via a proposed \textit{Generalized Focal Loss} function. Experiments demonstrate improved performance on rare and \textit{out-of-distribution} samples in autonomous driving datasets. Compared to traditional training and results (\textcolor{red}{red box}), our method trained with generalized focal loss can successfully detect rare classes like van and cyclist (\color{ForestGreen}{green box}).}

    \label{fig:teaser}
\end{figure*}

%% Introduction
\section{Introduction}
% \vspace{-0.1in}

Current methods for tackling data imbalance in machine learning either use some sort of \textit{Importance Weighting} \cite{importance-weighing}, or weigh the loss based on model prediction confidence for classification \cite{focal-loss}. The problem of dataset bias as understood in the literature for tasks such as object detection only refers to the imbalance in the class of objects such as \textit{car}, \textit{pedestrian}, \textit{cyclist}, or high level features like lighting and shadow conditions \cite{Zhang_2022_CVPR, krishnan2023lane}. They rely on class distribution, however do not consider perceptual feature likelihood as a way of understanding the data distribution. This is problematic, as class labels do not encapsulate fine-grained details like an object's context, occlusion, resolution, etc. Weighting by image likelihood captures these nuances missed by class information alone. This  a dataset that was collected 70\% in NYC and 30\% in the Mojave Desert with equal class distribution will still have a huge dataset imbalance, and without this information available as a part of dataset metadata, it will be impossible to account for with existing methods. Furthermore, our proposed method does not require a labeled dataset and rather operates on raw data samples, making it applicable to unsupervised techniques such as DINO \cite{caron2021emerging} for computer vision tasks.

Existing work in computer vision literature considers all samples to be equally important and defines objective functions without considering how likely it is to occur within a given dataset. If we instead weighed errors for less likely samples higher than more likely samples, it would encourage the model to put more attention on those samples, thereby `equalizing' the scales for all data samples.

% In this paper, we start by understanding the data distribution based on image appearances, and then utilize sample likelihood as a way to minimize dataset imbalance problems in computer vision.  This is achieved by first extracting image embeddings for each sample in a high-dimensional space using a pre-trained model, and then clustering them together to group similar-looking image frames. Relative sizes of these clusters then define the occurrence likelihood of each cluster sample in the training set.

% \LZ{Rewrite this part of workflow, break it down with 1. 2. 3.}

The proposed method tackles data imbalance via: (1) Image embeddings are extracted for each sample using a pre-trained model, mapping them to a high-dimensional feature space; (2) These embeddings are clustered together based on appearance similarity, grouping frames with similar visual characteristics; (3) The relative size of each cluster indicates the likelihood of samples in that cluster occurring in the dataset. These likelihoods are then utilized to reweight training losses. By first embedding frames into a perceptual feature space and clustering based on image semantics, we can estimate sample occurrence probabilities without relying on class labels. Reweighting the loss function by these computed likelihoods helps address imbalances in the visual data distribution.

The main contributions of this work are:
% \vspace{-0.1in}
\begin{enumerate}
    \vspace{-0.05in}
    \item A novel framework to prepare sample likelihoods information bank by clustering semantic embeddings of raw pixels. Unlike class-based techniques, this captures inter-object relationships and data heterogeneity missed by categorical labels alone.
    \vspace{-0.1in}
    \item Introduction of novel \textit{Generalized Focal Loss} that reweights by computed likelihoods, improving modeling of rare and OOD classes.
    \vspace{-0.1in}
    \item State-of-the-art results on major autonomous driving benchmarks, with over 200\% AP gains on rare classes like cyclist. 
    % \item Through quantitative analysis, we show that this method is able to surpass the state-of-the-art in camera-based 3D object detection, specifically for smaller datasets and under-represented objects. We demonstrate that, as the problem gets tougher, this method shortens the gap between the performance of the model on under-represented classes and the performance on the \textit{easy} classes, thereby equalizing the scales throughout the dataset.
    \vspace{-0.1in}
    \item Ablations demonstrating generalizability across datasets and complementarity to class-based techniques.
    \vspace{-0.1in}
    \item Open source code to enable further research into image-based likelihoods for mitigating dataset bias. By looking beyond class labels to raw pixels, this work opens promising new directions.
.
\end{enumerate}
% \vspace{-.2in}

%% Related Work
\section{Related Work}
% \vspace{-0.1in}
\textbf{Class-Based Resampling: } Class-based resampling involves either oversampling the under-represented classes or undersampling the over-represented classes in order to balance the data distribution \cite{10.1145/2907070}. This can be achieved by a variety of methods, such as generating synthetic data for the under-represented classes \cite{Jaipuria_2020_CVPR_Workshops}, randomly selecting a subset of the over-represented classes, or randomly duplicating examples in minority classes. However, this may be problematic and may increase the likelihood of overfitting \cite{5128907}. Region-Proposal-Network (RPN) \cite{NIPS2015_14bfa6bb} based architecture also suffers from class imbalance problem because of a large number of \textit{easy negatives}, and a common solution is to perform some form of \textit{hard negative mining} \cite{10.5555/929901, 990517, 5539906, journals/corr/ShrivastavaGG16}.

\textbf{Confidence-Based Weighing: } The idea behind confidence-based weighing is to encourage the model to put more emphasis on samples that are difficult to classify, thereby reducing the impact of data imbalance on the model's performance. It has been applied to a variety of tasks, such as object detection and image classification. One example of confidence-based weighing is focal loss \cite{focal-loss}, which modifies the standard cross-entropy loss function by adding a scaling factor that down-weights the loss for well-classified samples. This encourages the model to focus on difficult samples to classify, improving its performance on imbalanced data.

Other methods for confidence-based weighing have also been proposed, such as dynamically weighted balanced loss \cite{9324926} and class-balanced loss \cite{cui2019classbalancedloss, 9081913}. These methods have shown promising results for addressing data imbalance in various tasks.

\textbf{Image Embeddings for Dataset Analysis: } Image embeddings are a powerful tool for understanding the distribution of data in a dataset. They are a mathematical representation of an image in a high-dimensional space, where similar images are mapped to nearby points. Image embeddings can be used for dataset analysis by first extracting the embeddings for each image in the dataset using a pre-trained model. These embeddings can then be clustered together using unsupervised learning techniques, such as DBSCAN clustering, to group similar-looking images. The relative sizes of these clusters provide insight into the distribution of the data, allowing for the identification of any bias or imbalance in the dataset \cite{8658633, 9857264, 9093421}.

\textbf{Image-Based 3D Object Detection: } Recent advances in computer vision have led to the development of image-based 3D object detection methods, which use only camera images as input in autonomous driving. These methods are typically based on a Convolutional Neural Network (CNN) \cite{cubifae3d, park2021dd3d, simonelli2019disentangling} or a combination of CNN and Transformers \cite{li2022bevformer, wang2021detrd} to process the images and predict the 3D location and orientation of objects in the scene. For 3D bounding-box predictions, either the perspective view or a learned mapping from the perspective view to Bird’s Eye View (BEV) \cite{li2022bevformer, liu2022bevfusion, huang2021bevdet, zhang2022beverse} representation of the scene is used. Several methods have been proposed to improve further the accuracy of 3D object detection, including the use of depth information \cite{cubifae3d, park2021dd3d}, multiple cameras \cite{li2022bevformer, liu2022bevfusion, huang2021bevdet, zhang2022beverse}, temporal information \cite{li2022bevformer, zhang2022beverse}, and multiple modalities \cite{liu2022bevfusion, li2022deepfusion, pang2020clocs, shrivastava2021vr3dense}.

%% Methodology
\section{Methodology}
% \vspace{-0.1in}
The first step in enforcing dataset equity is understanding whether or not bias exists within the dataset based on sample appearance. Once we have determined that, some level of bias quantification is required to attempt combating this particular type of bias. 
In the following sections, we first build an intuition for this kind of dataset bias in autonomous driving datasets and then propose a way of quantifying it in Section \ref{sec:data_analysis}. The design of a loss function to tackle such dataset biases is further discussed in Section \ref{sec:gen_focal_loss} which allows us to boost performance on under-represented samples for camera-based 3D object detection tasks.

% \begin{figure}
%   \centering{\includegraphics[width=\linewidth]{fig/DatasetEquity_architecture.png}\label{fig:kitti-data-dist}}
%   % \vspace{-0.2in}
%   \caption{High-dimensional features extracted from images are first projected down onto a lower-dimensional space (e.g. 3D) using  a method such as \texttt{t-SNE}. These features are then clustered using an algorithm such as \texttt{DBSCAN} to identify frames with similar semantics in the same bucket. Relative sizes of these clusters define sample likelihoods, which are further used to compute \textit{Generalized Focal Loss} factor to weigh errors computed during the optimization process accordingly.}
% \end{figure}\label{fig:architecture}

\subsection{Dataset Description}
% \vspace{-0.1in}
In this work, two challenging autonomous driving datasets have been used to conduct our experiments with computer vision tasks: (1) nuScenes \cite{nuscenes}, and (2) KITTI \cite{Geiger2012CVPR} object detection dataset. We also analyze additional datasets such as WaymoOpenDataset \cite{Sun_2020_CVPR}, and BDD100K \cite{yu2020bdd100k}, to demonstrate further the type of biases that exists and attempts to quantify it.

\textbf{KITTI dataset:}\label{sec:kitti-dataset} 
KITTI 3D object detection benchmark is one of the most popular autonomous driving benchmarks and consists of $7481$ training samples, and $7518$ testing samples. KITTI dataset provides no validation set, however, it is common practice to split the training data into $3712$ training and $3769$ validation images as proposed in \cite{NIPS2015_6da37dd3}, and then report validation results. This benchmark consists of $8$ different classes but evaluates only $3$ classes: car, pedestrian, and cyclist. The evaluation metric is the  mean average precision (mAP) in 3D and Birds-Eye View (BEV) space, using a class-specific threshold on Intersection-over-Union (IoU). The average precision is computed using $41$ recall points (\texttt{$AP|_{40}$}) \cite{Simonelli2019DisentanglingM3}. The objects in the various splits are organized into three partitions according to their difficulty level (\textit{easy}, \textit{moderate}, \textit{hard}), and are evaluated separately.

\textbf{nuScenes dataset:}
nuScenes \cite{nuscenes}contains $1000$ driving scenes of 20s duration with keyframes annotated at 2Hz. Collected in Boston and Singapore, Collected across Boston and Singapore, it includes $28130$ training, $6019$ validation, and $6008$ test images from $6$ cameras. For 3D object detection, nuScenes provides 1.4M manually annotated boxes over $23$ classes. The official evaluation metric is nuScenes detection score (NDS), which aggregates mean average precision (mAP) and true positive errors for translation, scale, orientation, velocity, and attributes. NDS provides a holistic measure, unlike the mAP-based KITTI metric.

% This dataset also provides the official evaluation metrics for the 3D object detection task (\texttt{AP}) and is slightly different from the one used for \text{KITTI} dataset. In addition, the nuScenes metrics also contain 5 types of true positive ($\mathbb{TP}$) metrics, including \texttt{ATE}, \texttt{ASE}, \texttt{AOE}, \texttt{AVE}, and \texttt{AAE} for measuring translation, scale, orientation, velocity, and attribute errors, respectively. Lastly, this benchmark defines a nuScenes detection score (NDS) which captures all the errors and detection performance metrics.

\textbf{Waymo Open Dataset:}
Waymo Open Dataset \cite{Sun_2020_CVPR} has been introduced to the autonomous driving research community as a large-scale, high-quality, diverse dataset, which consists of $1150$ scenes that each span 20 seconds, across a range of conditions in multiple cities. The dataset is curated to resolve the lack of the environments variations. The geographical area covered in Waymo Open Dataset is larger than other autonmous driving datasets \cite{nuscenes} \cite{yu2020bdd100k} \cite{Geiger2012CVPR}, in terms of distribution of the coverage across geographies. 

\textbf{BDD100K dataset:}
BDD100K \cite{yu2020bdd100k} is a diverse driving video dataset with $100,000$ videos and annotation information for $10$ computer vision tasks. The dataset is designed to help train and evaluate models for autonomous driving applications. It includes a wide range of environments, including urban, and rural scenes, and covers various weather conditions, lighting conditions, and time of day. 

% \textbf{TuSimple dataset:} TuSimple dataset \cite{tusimple} is a large scale lane detection dataset consisting exclusively of highway driving scenes. It has 3626 clips for training, 2782 clips for testing and 358 clips for validation with each clip consisting on 20 frames. The dataset is collected in daylight under relatively stable weather conditions. We conduct data analysis on the whole dataset by combining train, validation and test splits. 

\subsection{Dataset Analysis}\label{sec:data_analysis}

In order to understand the data distribution, it is imperative to have some sense of similarity or distance between data samples. Image feature embeddings in a lower-dimensional space are extracted to compute the distance between two image frames. These features are then clustered together; the relative size of each cluster then gives us a sense of probabilities for samples within the given cluster occurring within the dataset.

\begin{figure}[h]
\centering
\includegraphics[width=0.35\textwidth]{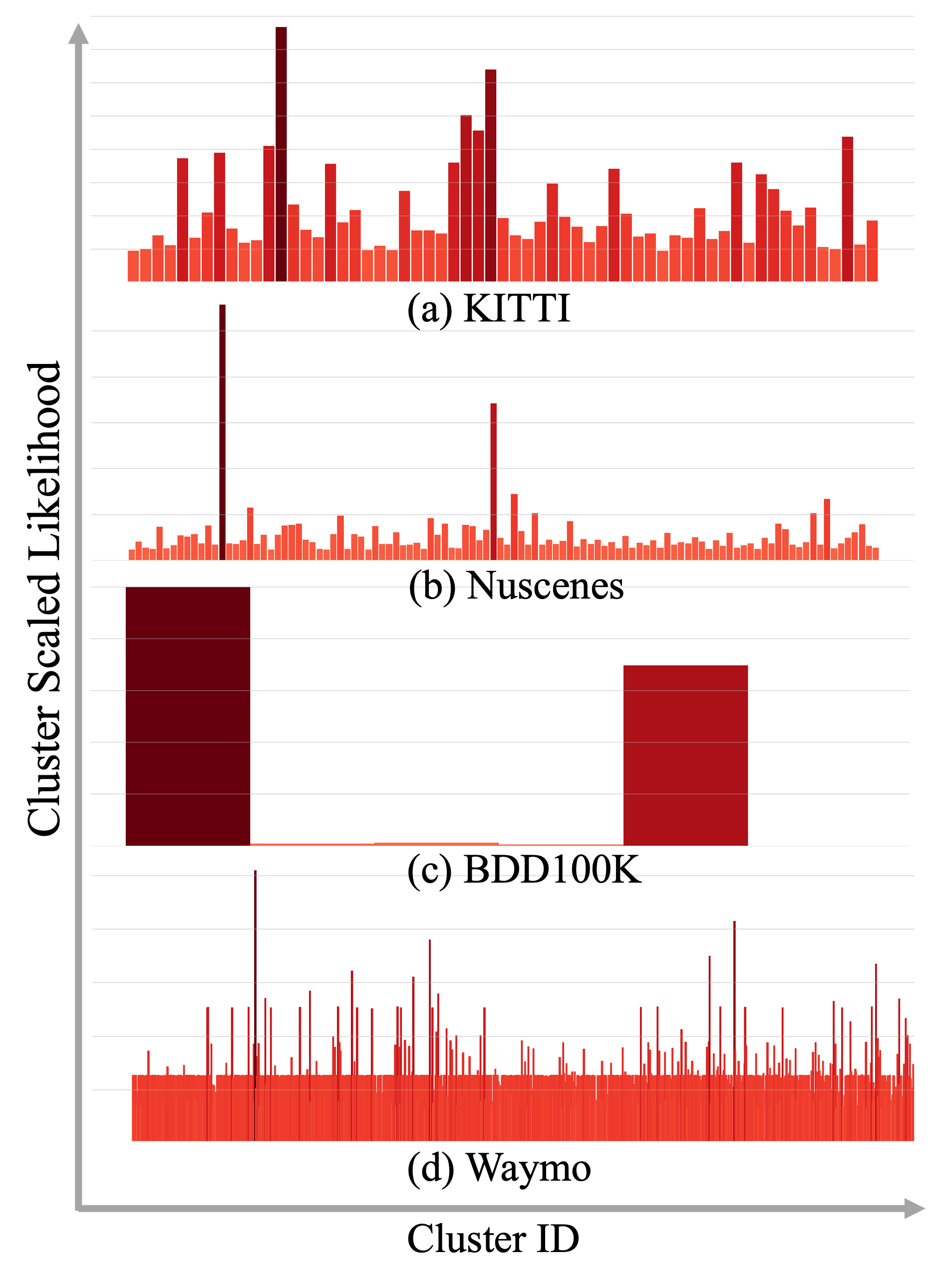}
\vspace{-0.1in}
\caption{To analyze dataset bias, this work extracts CLIP \cite{clip} embeddings of raw pixels and clusters them with HDBSCAN \cite{hdbscan} to discover non-uniform sample likelihoods. Unlike existing techniques relying on categorical labels, this captures visual relationships in an unsupervised manner. The derived likelihoods are then utilized to reweight loss functions, improving model performance on rare and out-of-distribution data. }
\label{fig:cluster-sampled-likelihood}
\vspace{-0.1in}
\end{figure}

% % \vspace{-0.1in}
% \begin{figure}[t]
% \includegraphics[width=0.43\textwidth]{fig/training_clusters/training_set.jpg}
% % \vspace{-0.1in}
% \caption{NuScenes, and KITTI dataset samples projected onto a 3-dimensional t-SNE space and then clustered using DBSCAN. 
% (only first 2-dimensions are visualized). Each color represents a unique cluster ID. For nuScenes, front camera was used, whereas, for KITTI dataset, left color stereo camera image was used.
% }
% \label{dataset_cluster}
% \end{figure}

Datasets considered in this work are: (1) nuScenes \cite{nuscenes}, and (2) KITTI object \cite{Geiger2012CVPR} (3) WaymoOpenDataset \cite{Sun_2020_CVPR}, and (4) BDD100K \cite{yu2020bdd100k}. For all of these datasets, each image sample from the front camera in the training set is passed through a pre-trained network such as ResNet101 \cite{resnet} or CLIP \cite{radford2021learning},
%and the output from the last hidden layer of dimension $2048$ is taken as 
to obtain the image embedding, which is a lower-dimensional vector summarizing image semantics and appearance.

Embedding from every image sample within the dataset is further projected onto a much lower-dimensional (3-dimensional) \texttt{t-SNE} \cite{tsne} space and then clustered together using algorithms like DBSCAN \cite{dbscan} or Hierarchical DBSCAN \cite{campello2013density}. \texttt{t-SNE} has shown to be effective for clustering applications, and as demonstrated by the authors of \cite{linderman2019clustering}, the early exaggeration phase can be leveraged as a powerful clustering tool that mimics the behavior of spectral clustering. Thus, in this paper, we use \texttt{t-SNE} as a pre-processing step before applying the clustering method to separate clusters in lower dimensions to both reduce the computation cost, and improve the clustering results.
% Figure \ref{fig:dataset_cluster} visualizes these clusters with a combination of ResNet101 and DBSCAN. 
% \textcolor{red}{In our experiments, the choice of a particular image feature extractor, or the clustering algorithm did not make much difference in terms of the final performance on downstream tasks.}

% \begin{figure}[!h]
%   \centering
%   \subfloat[nuScenes training dataset]
%   {\includegraphics[width=0.5\linewidth]{fig/nuscenes_training_clusters.png}
%   \label{fig:nuscenes_cluster}}
%   \subfloat[KITTI training dataset]
%   {\includegraphics[width=0.5\linewidth]{fig/kitti_training_clusters.png}
%   \label{fig:kitti_cluster}}
%   \caption{nuScenes and KITTI dataset samples projected onto a 3-dimensional t-SNE space and then clustered using DBSCAN (only first 2-dimensions are visualized). Each color represents a unique cluster ID. For nuScenes, front camera was used, whereas, for KITTI dataset, left color stereo camera image was used.}
% \end{figure}\label{fig:dataset_cluster}

Cluster size and the total number of samples in the training dataset are then used to compute the cluster likelihood of occurring within that dataset. This is a good representative of each sample likelihood within that cluster. This likelihood is computed as:

\begin{equation}\label{eqn:cluster_probabilities}
    P({C_i}) = \frac{|C_i|}{N}
\end{equation}

Where $C_i$ refers to the $i^{th}$ cluster, $|C_i|$ is the number of samples in $i^{th}$ cluster, and $N$ is the total number of samples within the training dataset.

These probabilities could be significantly low depending on the size of the training set. We, however, seek to compute probabilities relative to the largest cluster. This can be achieved by rewriting equation \ref{eqn:cluster_probabilities} as \ref{eqn:cluster_scaled_probabilities}, and then setting the scale factor $\tilde{K}=1.0$ to compute cluster likelihood as shown in equation \ref{eqn:cluster_scaled_likelihood}. This effectively gives us a relative likelihood of each cluster with respect to the largest cluster of data samples.
% \vspace{-0.05in}
\begin{equation}\label{eqn:cluster_scaled_probabilities}
    \begin{split}
    P({C_i}) &= \frac{|C_i|}{N} \frac{\max_i (|C_i|)}{\max_i (|C_i|)} = \frac{|C_i|}{\max_i (|C_i|)} \tilde{K}
    \end{split}
\end{equation}
% \vspace{-0.05in}
\begin{equation}\label{eqn:cluster_scaled_likelihood}
    \mathcal{L}_{s^{(i)}} = \frac{|C_i|}{\max_i (|C_i|)}
\end{equation}
% \vspace{-0.05in}
$\mathcal{L}_{s^{(i)}}$ in equation \ref{eqn:cluster_scaled_likelihood} is the scaled likelihood of samples corresponding to $i^{th}$ cluster. Histogram of these cluster likelihoods are shown in Figure \ref{fig:cluster-sampled-likelihood}.

% As observed in these plots, the samples' likelihood distribution is far from uniformly distributed and intrigues us to ask - \textbf{Are all samples created equal?}

% \begin{figure}[!h]
%   \centering
%   {\includegraphics[width=\linewidth]{fig/nuscenes_dataset_likelihood.png}}
%   \caption{nuScenes dataset samples cluster scaled probabilities}
% \label{fig:nuscenes_dataset_likelihood}
% \end{figure}

% \begin{figure}[!h]
%   \centering
%   {\includegraphics[width=\linewidth]{fig/kitti_dataset_likelihood.png}}
%   \caption{KITTI dataset samples cluster scaled probabilities}
% \label{fig:kitti_dataset_likelihood}
% \end{figure}

\subsection{Generalized Focal Loss}\label{sec:gen_focal_loss}

\begin{figure}[!h]
\centering
\includegraphics[width=0.4\textwidth]{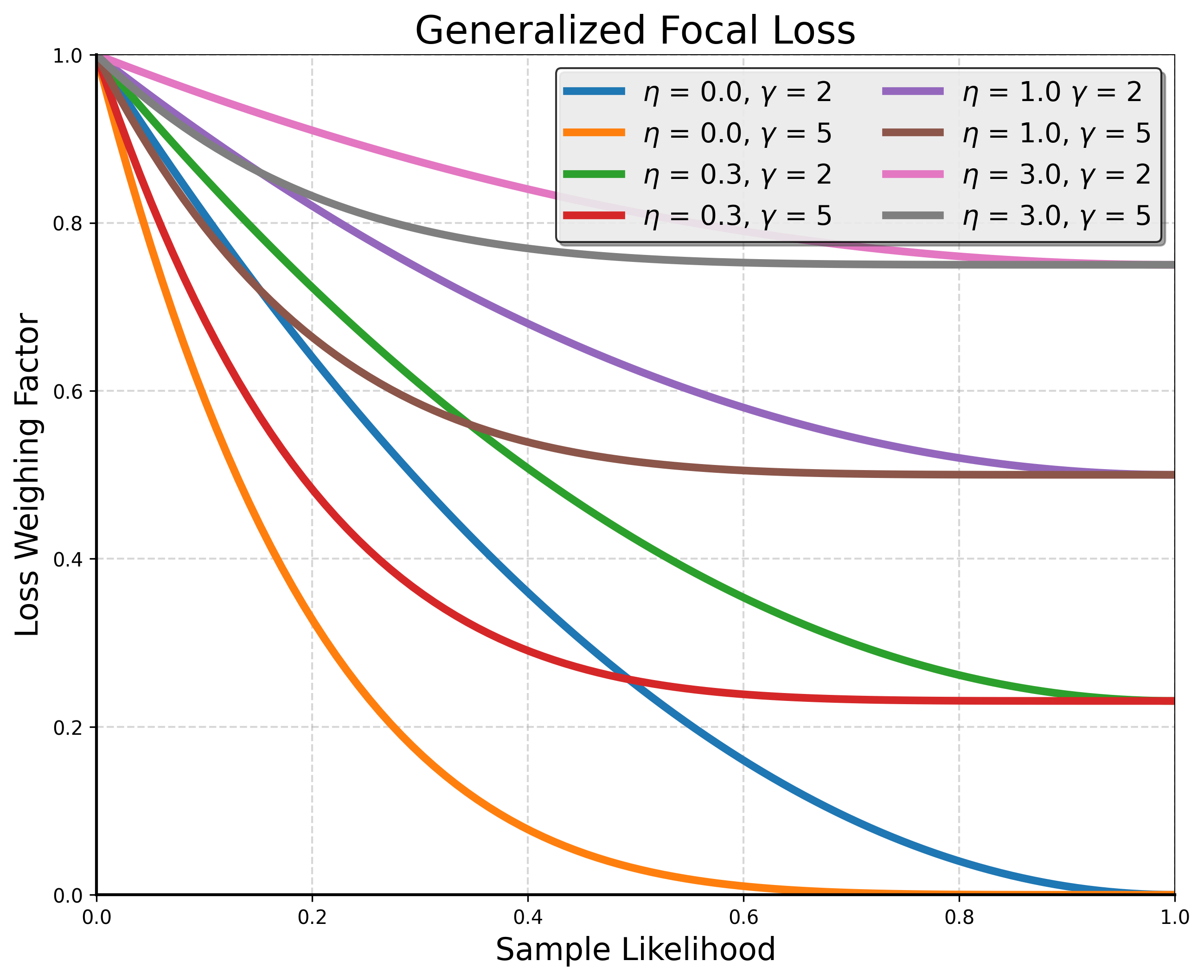}
\caption{We propose a novel loss function called \texttt{Generalized Focal Loss}, which addresses the issue of data imbalance in computer vision by weighting each sample differently based on its likelihood of occurrence, leading to improved performance on downstream computer vision tasks. Plot of this function for various $\eta$ and $\gamma$ is shown in this figure; note that it defaults to \texttt{Focal Loss} for $\eta=0.0$.}
\label{fig:gen_focal_loss}
\end{figure}

Knowledge of sample likelihoods allows us to weigh each sample loss differently, thereby enforcing \textit{dataset equity}. A new loss weighting function called \texttt{Generalized Focal Loss} as given in equation \ref{eqn:gen_focal_loss} is designed, which helps push the losses for less-likely samples high, and vice versa. Plots for this function against sample likelihood for various $\eta$ and $\gamma$ are shown in Figure \ref{fig:gen_focal_loss}. Note that, for $\eta=0$, this function defaults to \textit{Focal Loss} \cite{focal-loss} weighting function. 
% \vspace{-0.05in}
\begin{equation}\label{eqn:gen_focal_loss}
    \mathbf{W}_{gfl}(p, \eta, \gamma) = \frac{\eta + (1 - p)^\gamma}{\eta + 1}
\end{equation}
% \vspace{-0.05in}
This way of training neural networks for computer vision tasks is described in Figure \ref{fig:teaser}. Please note that we do not propose to replace \texttt{Focal Loss} with our loss, but rather to use them in conjunction, and we demonstrate these added benefits on baseline methods - all of which already use focal loss during the training process.

\subsection{Camera-Based 3D Object Detection}

This work takes camera-based 3d object detection as an example task to demonstrate the effectiveness of our proposed method. To test our method's generalizability, we conduct experiments on 3 existing methods, namely, BEVFormer \cite{li2022bevformer}, DD3D \cite{park2021dd3d}, and BEVFusion \cite{liu2022bevfusion}, across two datasets - nuScenes, and KITTI. These two methods are chosen because they are state-of-the-art for camera-based 3d object detection tasks on the two datasets and have their implementations publicly available. During the training, \texttt{Generalized Focal Loss} weight is computed for each sample and is multiplied to the total loss before backward propagation. This does not increase the number of parameters of the model, and only adds a single $multiplication$ node in the computational graph during backpropagation without affecting the inference time of the underlying tasks. This simple method of sample reweighing improves the \text{mAP} score of DD3D by over $200\%$ for under-represented classes of \text{KITTI} dataset, the \text{mAP} score of BEVFormer by $1.44\%$ on \text{nuScenes} dataset, and the \text{mAP} score of BevFusion camera-only model by $1.05\%$ on \text{nuScenes} dataset. Note that although applying on same nuscenes dataset, BEVFormer and BevFusion share different model architecture, and our method achieves consistent improvements over both models.

\begin{figure}[!h]
  \centering
  \subfloat{\includegraphics[width=0.5\linewidth]{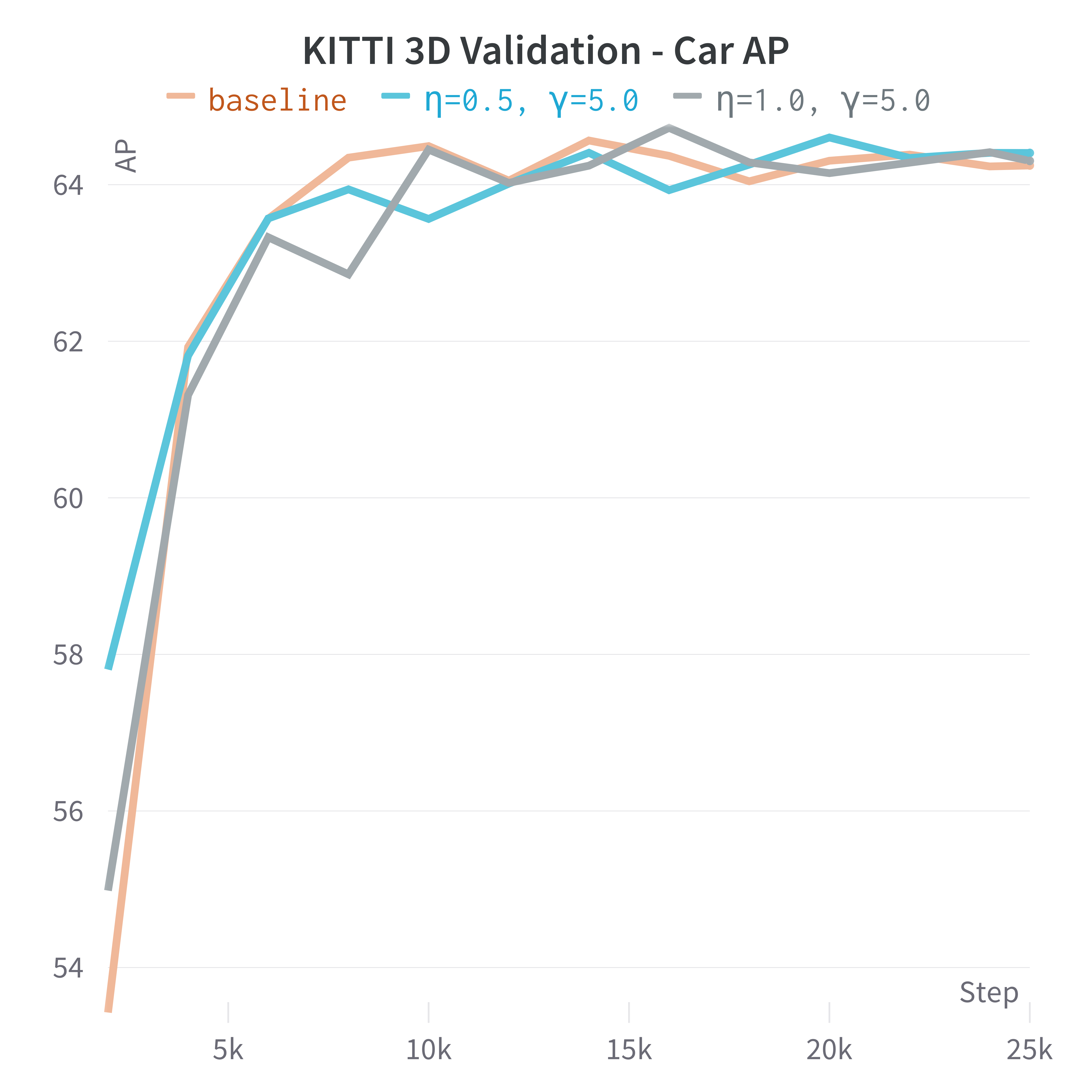}\label{fig:AP-training-car}}
  \centering
  \subfloat{\includegraphics[width=0.5\linewidth]{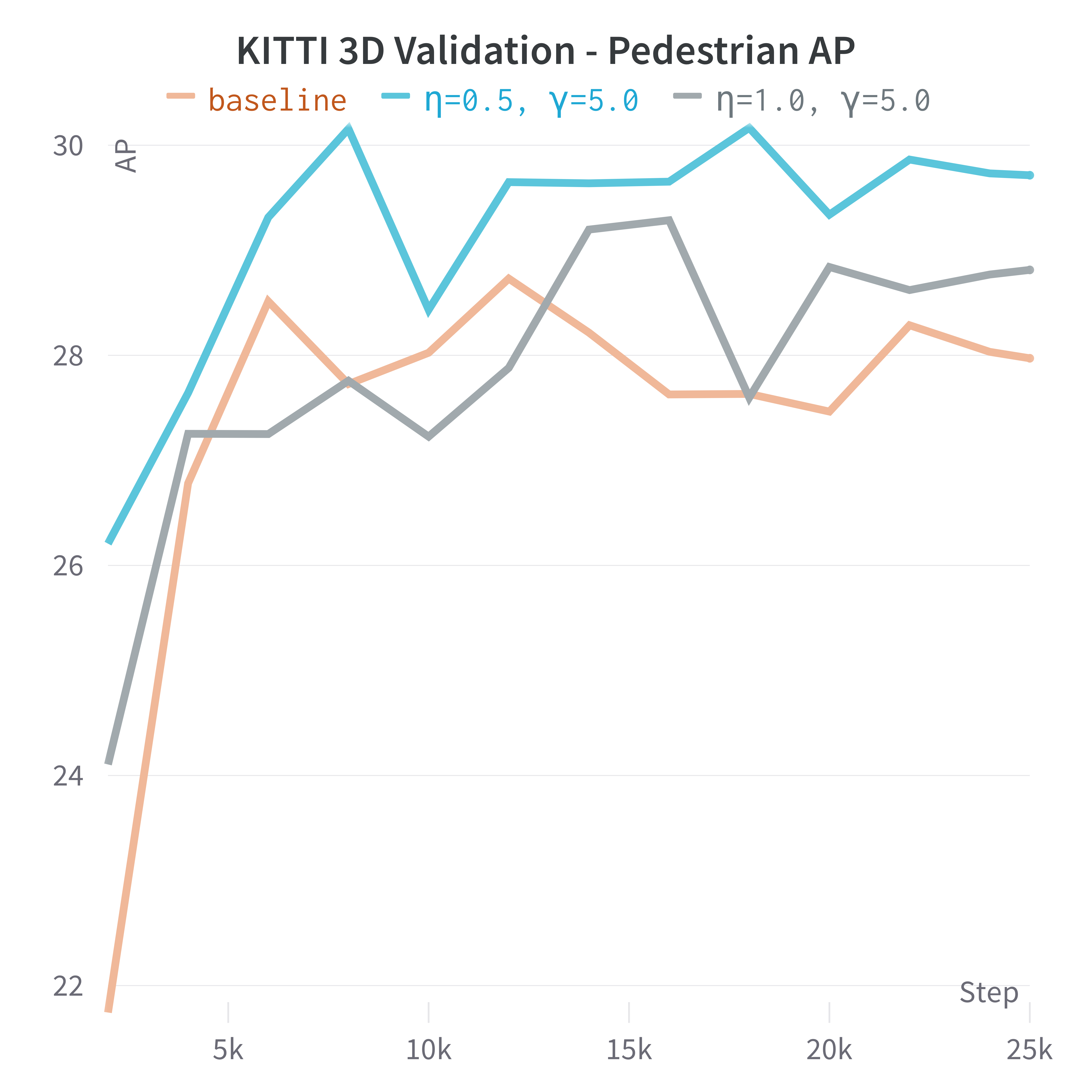}\label{fig:AP-training-ped}}
  \caption{Validation AP for \textit{car} and \textit{pedestrian} during training as a function of number of optimization steps. Model trained with the \textit{Generalized Focal Loss} achieves higher AP quickly on under-represented object classes, such as \textit{pedestrian} in the KITTI 3D object dataset.}
  \label{fig:AP-training-graph}
\end{figure}

Improvements with our methods are truly apparent for samples with low-likelihood, such as a camera frame containing object classes with very few instances within dataset, or common objects with partial occlusion/rare appearance. This is demonstrated in Figure \ref{fig:dd3d-qual-comparision}. 
Comparison of validation graphs during training also hints at the advantage of using our loss function on under-represented samples. While there does not seem to be much advantage for a well-represented object class such as \textit{car}, on \textit{pedestrian} classes the model seems to be converging quickly while achieving higher \texttt{AP} as shown in Figure \ref{fig:AP-training-graph}.

For BEVFormer, we train the \texttt{BEVFormer-base} and \texttt{BEVFormer-small} variants on the \textit{nuScenes} training dataset, and test it on both, $validation$, as well as $test$ split. DD3D, which is state-of-the-art for monocular 3D object detection on KITTI dataset, has been trained in two different modes, (1) the model is trained on the \texttt{train} split and tested on \texttt{val} split, and (2) the model is trained on the \texttt{trainval} split and tested on the \texttt{test} set on KITTI server. For BEVFusion baseline, the camera-only model has been trained from scratch on the \textit{nuScenes} training dataset, and validated on the nuScenes val dataset. More details about the experiments are given in Section \ref{sec:experiments}.

%% Experiments
\vspace{-0.1in}
\section{Experimental Section}\label{sec:experiments}
\begin{figure}[t]
\centering
\includegraphics[width=0.5\textwidth]{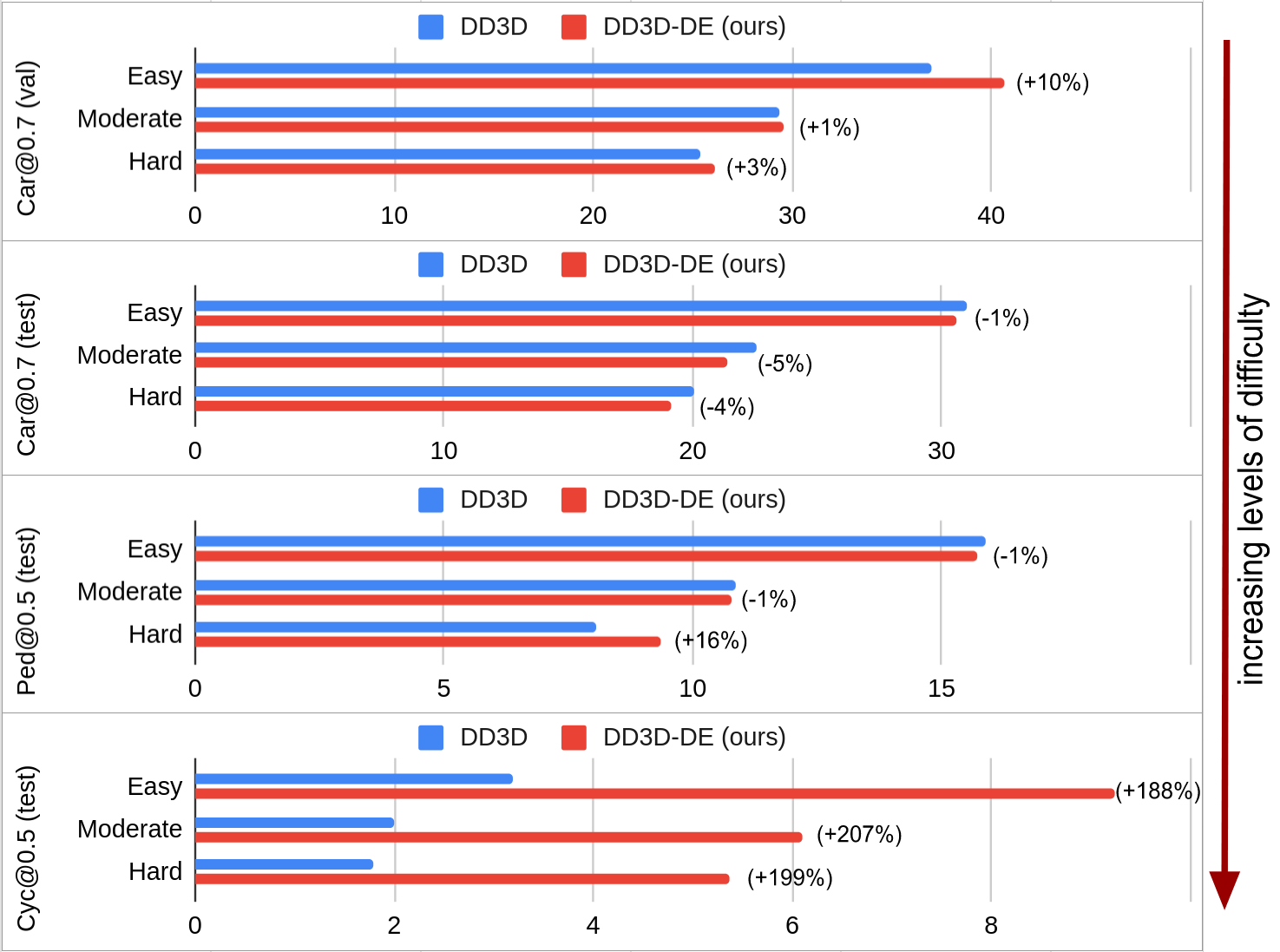}
\caption{Comparison of BEV AP for a camera-based 3D object detection method called DD3D, with and without the proposed \texttt{Generalized Focal Loss} function, showing a significant improvement in performance when using the proposed loss function, particularly as the difficulty of the problem increases. Levels of difficulty here refer to the object class \textit{rareness}, and within each object class, \textit{easy}, \textit{moderate}, and \textit{hard} add another level of hierarchy. The use of the \texttt{Generalized Focal Loss} function brings equity within datasets and equalizes scales for all samples.}
\end{figure}

To further validate the effectiveness of our loss function on different model architectures, three methods of camera-based 3D object detection on two different AV datasets have been benchmarked. While our method only improve performance of BEVFormer and BEVFusion models on the nuScenes benchmarking dataset by only about $1\%$, the improvement on KITTI dataset is substantial (~$10\%$ improvement on \textit{val} set), especially for the under-represented classes such as \textit{pedestrian}, and \textit{cyclist}, achieving over $200\%$ gain in mAP. This could be due to size of KITTI dataset is much smaller compared to the nuScene dataset and has a huge discrepancy in terms of class distribution in Table \ref{tab:kitti-dist}. 

In addition, we suspect that existing dataset balancing technique such as Class-balanced Grouping and Sampling(CBGS)\cite{zhu2019classbalanced} has played critical role on models trained with nuscenes, thus further experiments has been conducted in Table \ref{tab:nuscenes-val-bevfusion} that excludes the impact of CBGS, and ~$10\%$ improvement in NDS has been observed on our BEVFusion model. This shows the impact of considering features of a whole scene while sampling/weight balancing compared to label based sampling.
For all datasets, features extracted from samples are first projected down onto a $3D$ \texttt{t-SNE}\cite{tsne} space initialized with \texttt{PCA}\cite{Jolliffe:1986}, which is then followed by either \texttt{DBSCAN}\cite{dbscan}, or \texttt{HDBSCAN}\cite{campello2013density} algorithm for feature clustering to quantify sample likelihoods. Parameters used for clustering were $[\epsilon=2.0, \texttt{min\_samples=10}]$ for KITTI, $[\epsilon=1.0, \texttt{min\_samples}=10]$ for nuScenes, $[\epsilon=1.0, \texttt{min\_samples=10}]$ for WaymoOpenDataset, and $[\epsilon=1.5, \texttt{min\_samples=100}]$ for BDD100K.

% \vspace{-0.2in}
\begin{table}[!h]
\centering
\caption{Distribution of class instances in KITTI object \textit{training} dataset across 5 categories.}
\setlength\tabcolsep{3.5pt}
\begin{tabular}{l | c c c c c}
\toprule
Category & Car & Pedestrian & Van & Cyclist & Truck \\
\midrule
\rowcolor{gray95}
\# Instances & 28742 & 4487 & 2914 & 1627 & 1094 \\
\bottomrule
\end{tabular}
\label{tab:kitti-dist}
\end{table}

\begin{figure*}[!h]
  \centering
  \subfloat[\texttt{DD3D} incorrectly classifies a \textit{Van} as \textit{Car}]{\includegraphics[width=0.4\linewidth]{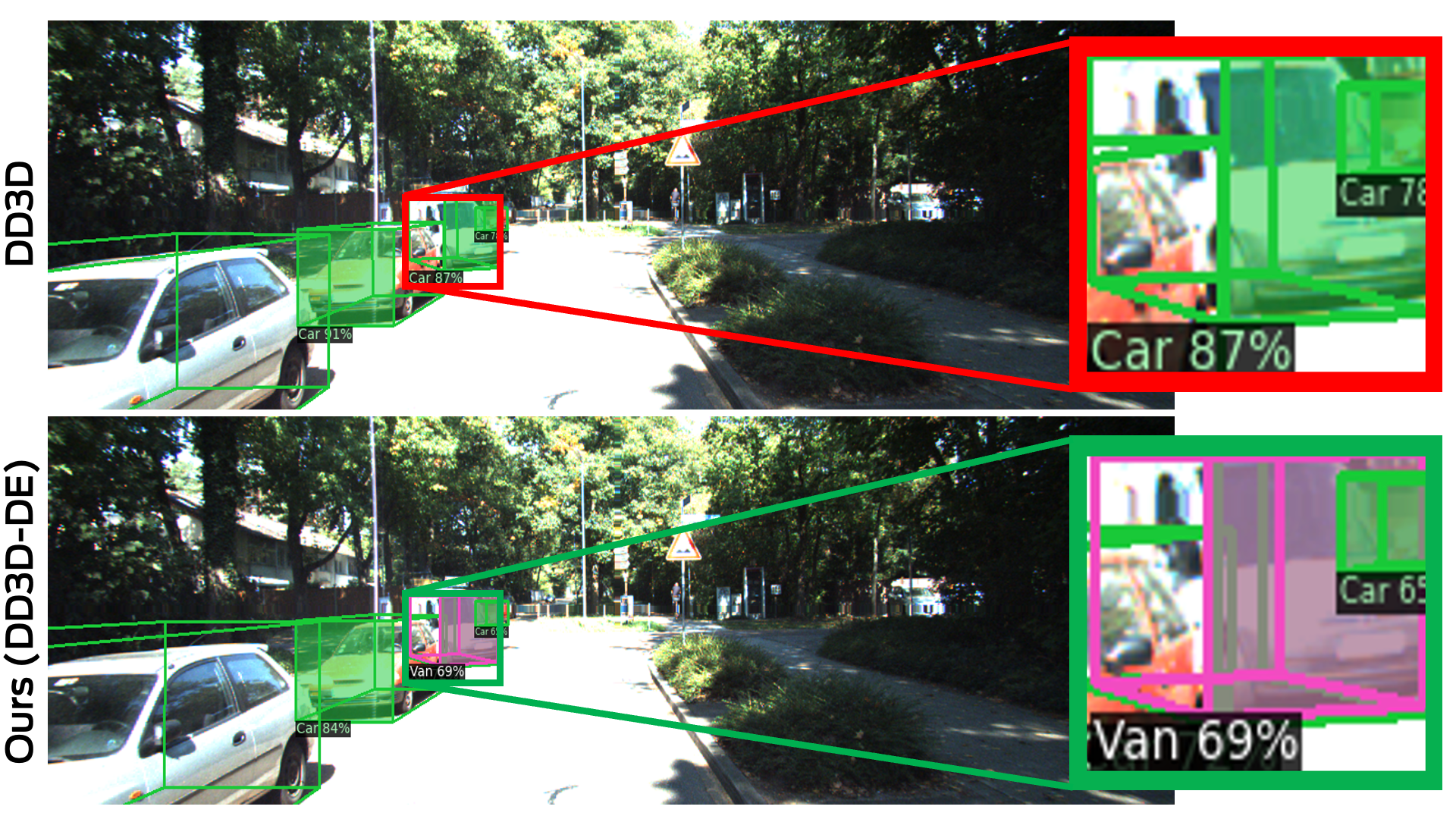}}
  \centering
  \subfloat[\texttt{DD3D} fails to detect a \textit{Van} and a \textit{Cyclist}]{\includegraphics[width=0.4\linewidth]{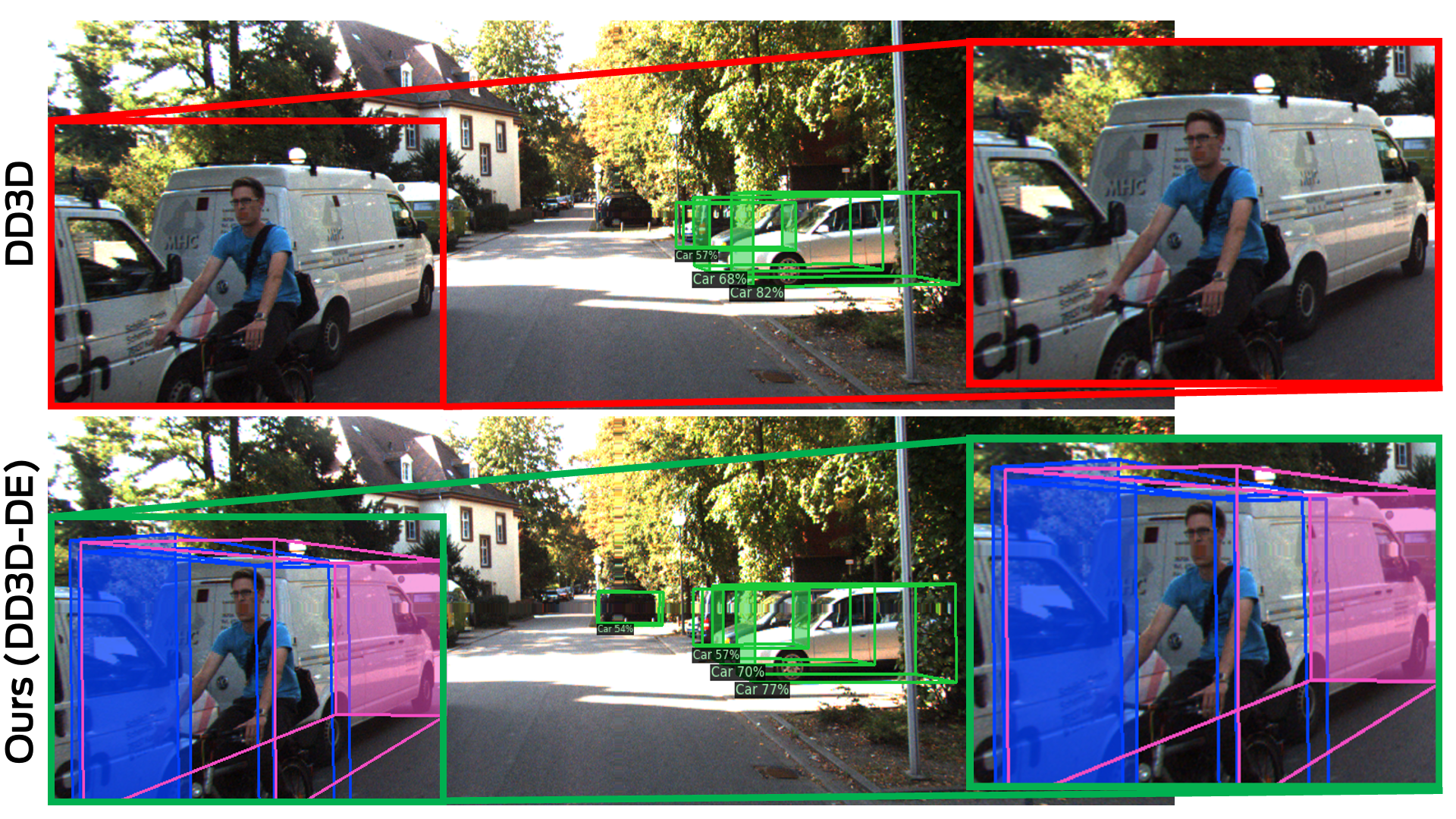}}

  \centering
  \subfloat[\centering{\texttt{DD3D} fails to detect a \textit{Cyclist}; \texttt{DD3D-DE} succeeds in detecting, but incorrectly classifies as \textit{Pedestrian}}]{\includegraphics[width=0.4\linewidth]{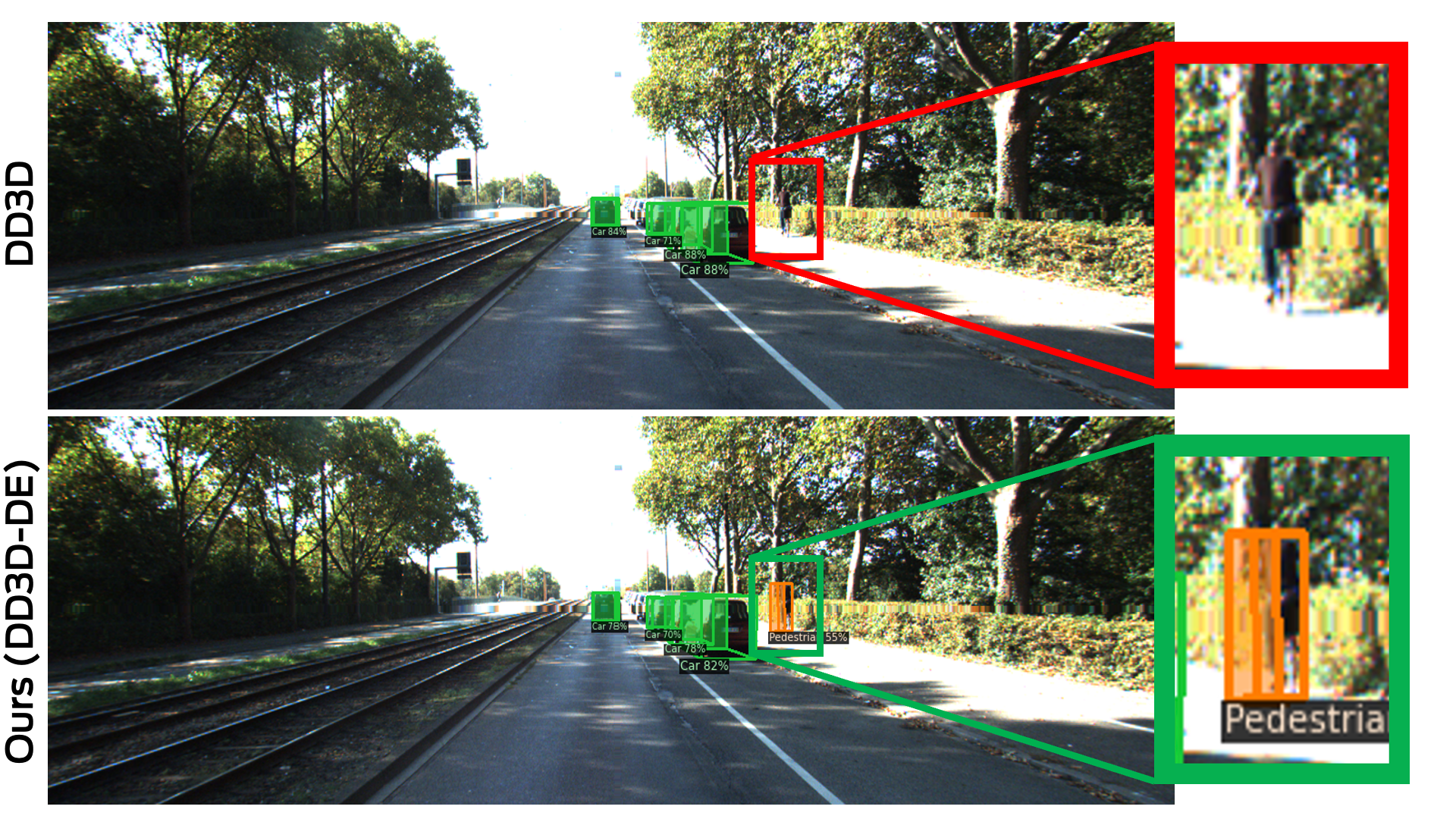}}
  \centering
  \subfloat[\centering{\texttt{DD3D} fails to detect partially occluded car represented by a very few number of pixels}]{\includegraphics[width=0.4\linewidth]{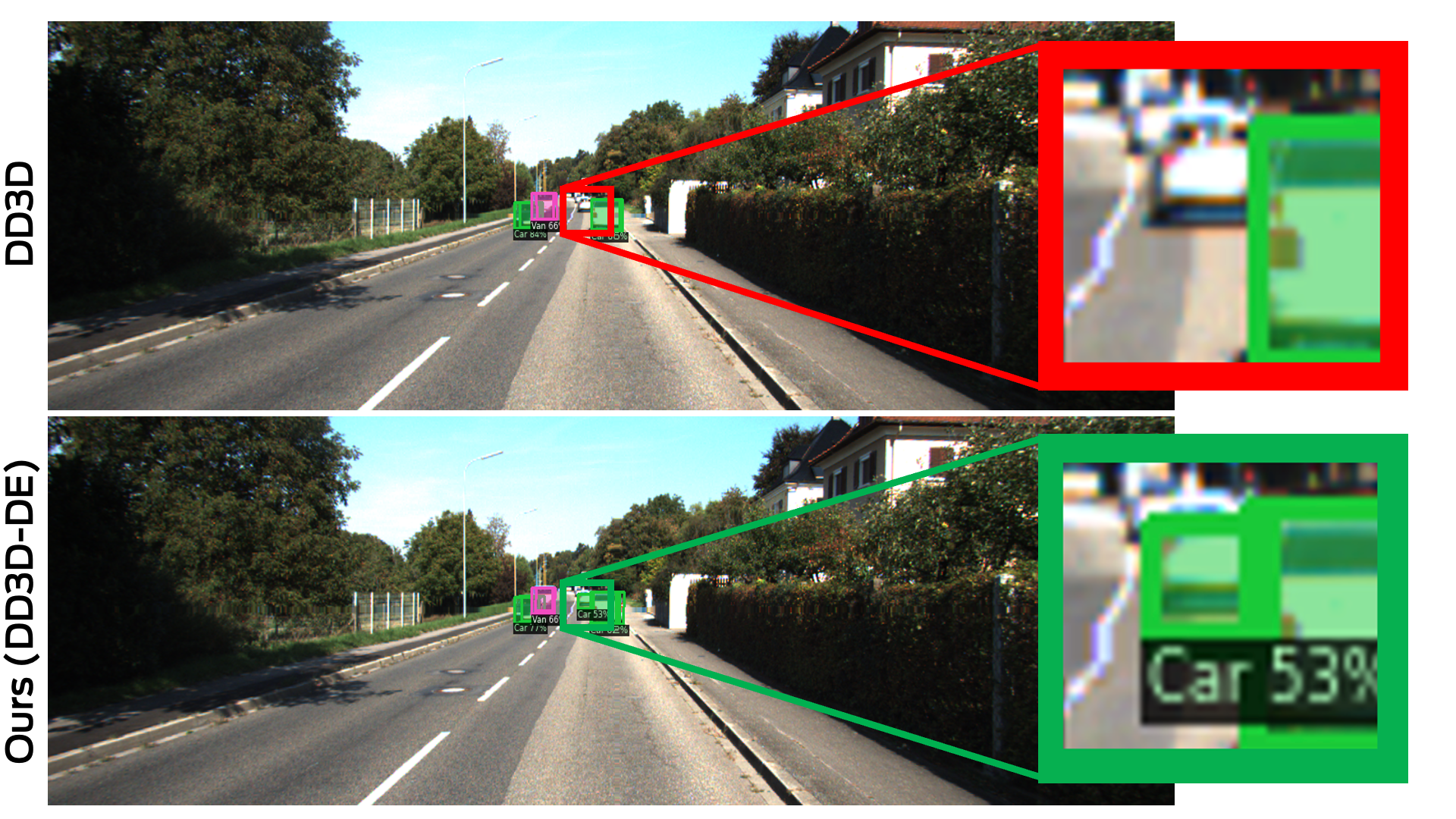}}

  \centering
  \subfloat[\centering{Orientation predicted by \texttt{DD3D} for a \textit{Car} far away is highly offset}]{\includegraphics[width=0.4\linewidth]{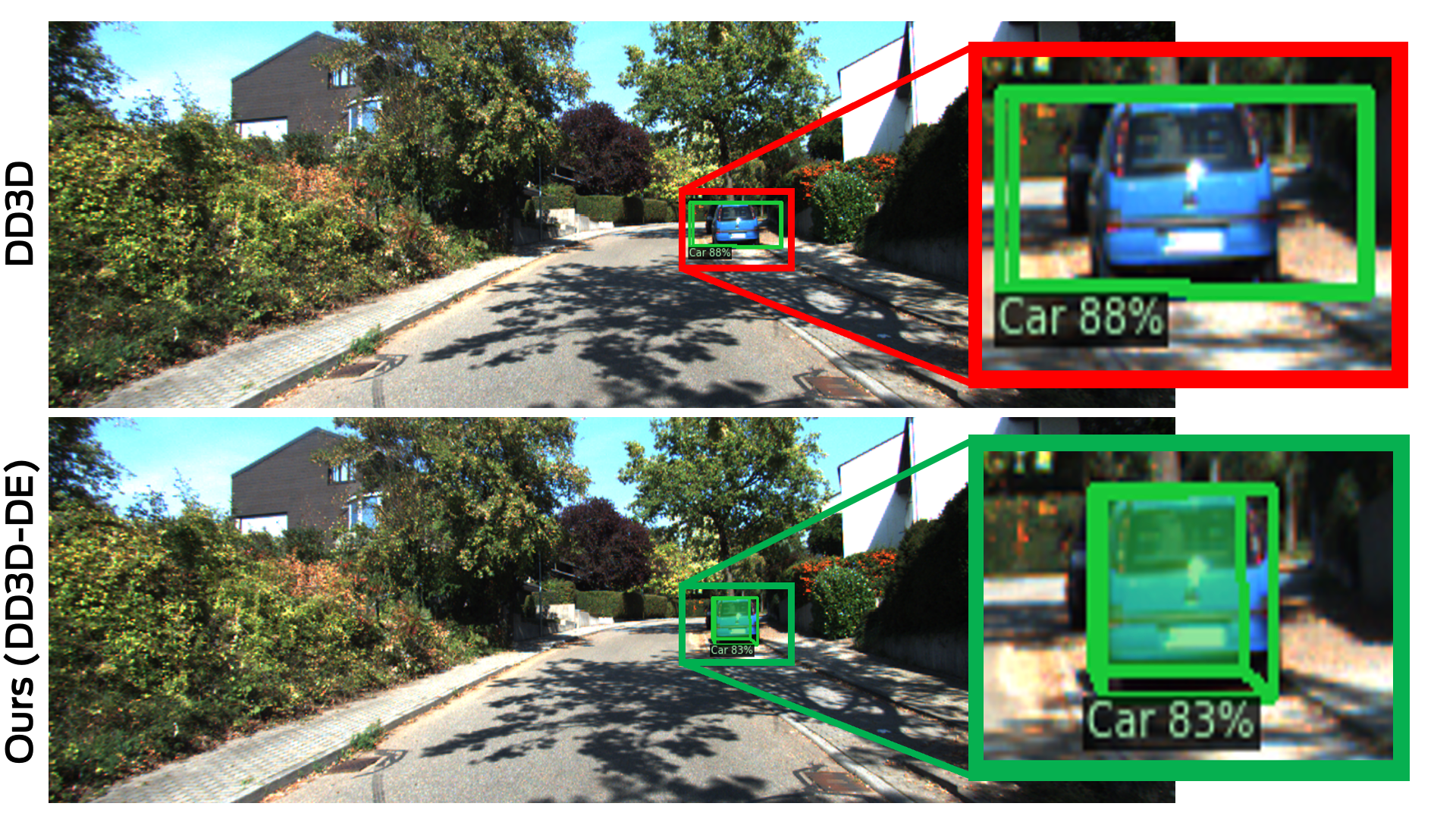}}
  \centering
  \subfloat[\centering{\texttt{DD3D} misses to detect a \textit{Car} at a farther distance which appearance easily blends with the background}]{\includegraphics[width=0.4\linewidth]{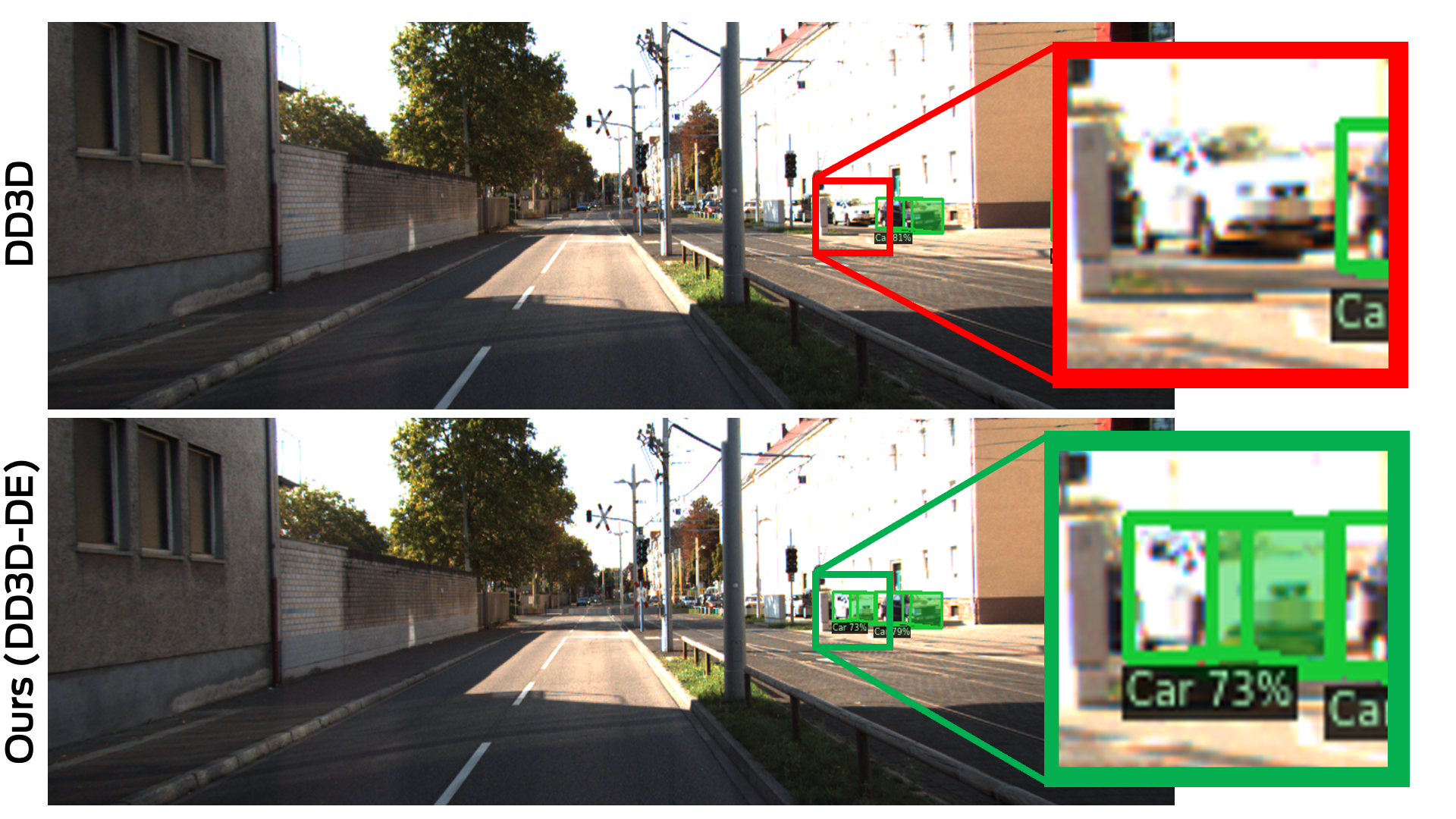}}
  
  \caption{Qualitative analysis of predictions from the baseline \texttt{DD3D} model and our \texttt{DD3D-DE} model. The samples shown here were randomly drawn from the split of the KITTI \textit{test} split. As shown in the images, \texttt{DD3D-DE} improves the performance over the baseline model on under-represented, and out-of-distribution samples containing objects such as \textit{Van}, \textit{Cyclist}, and occluded or far away \textit{Car}}.
  \vspace{-0.2in}
  \label{fig:dd3d-qual-comparision}
\end{figure*}

\textbf{DD3D:} V2-99 \cite{lee2019centermask} extended to an FPN was used to train the DD3D model on KITTI dataset. Quantitative results of the model trained with our loss function on \textit{val} set and \textit{test} set are shown in Tables \ref{tab:kitti-val} and \ref{tab:kitti-test} respectively. We split the data into \textit{train} and \textit{val} set as described in \ref{sec:kitti-dataset}, train the model on \textit{train} set, and report the results on \textit{val} set. For reporting the results on \textit{test} split, we train on the entire \textit{trainval} split, and then submit our predictions on \textit{test} split to the KITTI server for benchmarking. Qualitative analysis of the improvements brought by our method over the baseline \texttt{DD3D} is demonstrated in Figure \ref{fig:dd3d-qual-comparision}, which clearly illustrates how the \textit{Generalized Focal Loss} really shines for methods working with datasets where all samples might not be equally likely. These models were trained for a total of $25000$ steps with a batch size of $24$, on a single NVIDIA A100 GPU. Initial learning rate of $0.002$, with \textit{MultiStepLR} scheduler was used during training \footnote{DD3D official implementation: \url{https://github.com/TRI-ML/dd3d}}.

\begin{table}[!t]
\centering
\caption{\textbf{3D detection results on KITTI \texttt{val} set.} The suffix \texttt{DE} signifies our method of applying \textit{Generalized Focal Loss} weights to each sample. The best results are highlighted in \textbf{bold}. \texttt{Class@N} in this table refers to the $AP|_{R40}$ score computed for Class at an IoU threshold of $N$.}
\begin{tabular}{l c c | c c c }
\toprule
Methods & $\eta$ & $\gamma$ & \multicolumn{3}{c}{Car@0.7} \\
\midrule
\rowcolor{gray95}
 & & &\multicolumn{3}{c}{BEV AP} \\
\rowcolor{gray95}
 & & & Easy & Mod & Hard \\
\midrule
DD3D & - & - & 37 & 29.4 & 25.4 \\
\rowcolor{gray95}
DD3D-DE & 0.5 & 5.0 & 39.381 & 29.591 & 25.868 \\
\rowcolor{gray9}
DD3D-DE & 1.0 & 5.0 & \textbf{40.607} & \textbf{29.598} & \textbf{26.087} \\

\midrule
$\Delta$ & & &+9.75\% & +0.67\% & +2.70\% \\
\bottomrule
\end{tabular}
\label{tab:kitti-val}
\vspace{-0.2in}
\end{table}
\raggedbottom

\begin{table*}[]
\centering
\caption{\textbf{3D detection results on KITTI \texttt{test} set.} The suffix \texttt{DE} signifies our method of applying \textit{Generalized Focal Loss} weights to each sample. Best results are highlighted in \textbf{bold}. Value of $\eta$ and $\gamma$ in the \textit{Generalized Focal Loss} weight was set to $1.0$, and $5.0$ respectively. \texttt{Class@N} in this table refers to the $AP|_{R40}$ score computed for Class at an IoU threshold of $N$.}
\setlength\tabcolsep{1.3pt}
\begin{tabular}{l | c c c | c c c | c c c | c c c | c c c | c c c }
\toprule
Methods & \multicolumn{6}{c|}{Car@0.7} &  \multicolumn{6}{c|}{Pedestrian@0.5} &  \multicolumn{6}{c}{Cyclist@0.5} \\
\midrule
\rowcolor{gray95}
 & \multicolumn{3}{c}{BEV AP} & \multicolumn{3}{c|}{3D AP} & \multicolumn{3}{c}{BEV AP} & \multicolumn{3}{c|}{3D AP} & \multicolumn{3}{c}{BEV AP} & \multicolumn{3}{c}{3D AP} \\
\rowcolor{gray95}
 & Easy & Mod & Hard & Easy & Mod & Hard & Easy & Mod & Hard & Easy & Mod & Hard & Easy & Mod & Hard & Easy & Mod & Hard \\
\midrule
M3D-RPN\cite{brazil2019m3drpn} & 21.02 & 13.67 & 10.23 & 14.76 & 9.71 & 7.42 & 5.65 & 4.05 & 3.29 & 4.92 & 3.48 & 2.94 & 1.25 & 0.81 & 0.78 & 0.94 & 0.65 & 0.47 \\
MonoDIS\cite{9200697} & 24.45 & 19.25 & 16.87 & 16.54 & 12.97 & 11.04 & 7.79 & 5.14 & 4.42 & 9.07 & 5.81 & 5.09 & 1.17 & 0.54 & 0.48 & 1.47 & 0.85 & 0.61 \\
\midrule
DD3D & \textbf{30.98} & \textbf{22.56} & \textbf{20.03} & \textbf{23.22} & \textbf{16.34} & \textbf{14.2} & \textbf{15.9} & \textbf{10.85} & 8.05 & 13.91 & 9.3 & 8.05 & 3.2 & 1.99 & 1.79 & 2.39 & 1.52 & 1.31 \\
\rowcolor{gray9}
DD3D-DE & 30.62 & 21.37 & 19.1 & 21.71 & 14.76 & 12.92 & 15.7 & 10.77 & \textbf{9.35} & \textbf{14.28} & \textbf{9.57} &\textbf{8.2} & \textbf{9.23} & \textbf{6.11} & \textbf{5.36} & \textbf{7.23} & \textbf{4.61} & \textbf{4.1} \\

\midrule
\% $\Delta$ & -1 & -5 & -4 & -6 & -9 & -9 & -1	& -1 & +16 & +3 & +3 & +2 & +188 & +207 & +199 & +203 & +203 & +213\\

\midrule
\multicolumn{19}{l}{As the task gets tougher, \texttt{DD3D-DE} performs better. $\longmapsto$} \\
\midrule
\bottomrule
\end{tabular}
\label{tab:kitti-test}
\end{table*}
\raggedbottom

\textbf{BEVFormer:} Experiments with BEVFormer was done on with two different variants of the network architectures: (1) \texttt{BEVFormer-base}, and (2) \texttt{BEVFormer-small}. Both architectures used ResNet-101 \cite{7780459} as the backbone network. With the official codebase \footnote{BEVFormer official implementation: \url{https://github.com/fundamentalvision/BEVFormer}}, we were unable to reproduce the results reported in the paper, and hence, the baseline metrics reported in this paper were computed by retraining the models from scratch. These results are summarized in Tables \ref{tab:nuscenes-test} and \ref{tab:nuscenes-val}.

\begin{table*}[!h]
\centering
\caption{\textbf{3D detection results with DD3D on nuScenes \texttt{test} set.} The suffix \texttt{DE} signifies our method of applying \textit{Generalized Focal Loss} weights to each sample. R101 in the table is refers to the ResNet-101 backbone. Best NDS and mAP scores are highlighted in \textbf{bold}.}
\setlength\tabcolsep{2.2pt}
\begin{tabular}{l c c c | c c | c c | c c c c c c }
\toprule
Method & Backbone & $\eta$ & $\gamma$ & NDS$\uparrow$ & $\Delta$ & mAP$\uparrow$ & $\Delta$  & mATE$\downarrow$     & mASE$\downarrow$     & mAOE$\downarrow$     & mAVE$\downarrow$    & mAAE$\downarrow$    \\
\midrule
FCOS3D \cite{wang2021fcos3d} & R101 & - & - & 0.428 & - & 0.358 & - & 0.690 & 0.249 & 0.452 & 1.434 & 0.124 \\
PGD \cite{wang2021probabilistic} & R101 & - & - & 0.448 & - & 0.386 & - & 0.626 & 0.245 & 0.451 & 1.509 & 0.127 \\

\midrule
\rowcolor{gray95}
BEVFormer-base & R101 & - & - & 0.5196 & - & 0.4242 & - & 0.6351 & 0.2684 & 0.4219 & 0.4593 & 0.1406 \\ 
\rowcolor{gray9}
BEVFormer-base-DE & R101 & 3.0 & 5.0 & \textbf{0.5198} & (0.04\%) & \textbf{0.4303} & (1.44\%) & 0.6369 & 0.2706 & 0.4418 & 0.4649 & 0.1388 \\ 

\bottomrule
\end{tabular}
\label{tab:nuscenes-test}
\vspace{-0.1in}
\end{table*}

\begin{table*}[!h]
\centering
\caption{\textbf{3D detection results with BEVFormer on nuScenes \texttt{val} set.} BEVFormer-sml stands for the \texttt{BEVFormer-small} variant, and the suffix \texttt{DE} signifies our method of applying \textit{Generalized Focal Loss} weights to each sample. R101 in the table is refers to the ResNet-101 backbone. Best NDS and mAP scores are highlighted in \textbf{bold}.}
\setlength\tabcolsep{2.2pt}
\begin{tabular}{l c c c | c c | c c | c c c c c c }
\toprule
Method & Backbone & $\eta$ & $\gamma$ & NDS$\uparrow$ & $\Delta$ & mAP$\uparrow$ & $\Delta$  & mATE$\downarrow$     & mASE$\downarrow$     & mAOE$\downarrow$     & mAVE$\downarrow$    & mAAE$\downarrow$    \\
\midrule

\rowcolor{gray95}
BEVFormer-sml & R101 & - & - & 0.3874 & - & 0.3516 & - & 0.787 & 0.2885 & 0.4981 & 1.169 & 0.3101 \\
\rowcolor{gray9}
% BEVFormer-sml-DE & R101  & 0.5 & 5.0 & 0.3886 & 0.3509 & 0.7984 & 0.2932 & 0.4878 & 1.109 & 0.2892 \\
\rowcolor{gray9}
BEVFormer-sml-DE & R101  & 0.5 & 2.0 & 0.3907 & (0.85\%) & 0.3544 & (0.80\%) & 0.7687 & 0.2904 & 0.4952 & 1.174 & 0.3104 \\
\rowcolor{gray9}
BEVFormer-sml-DE & R101  & 1.0 & 2.0 & 0.39 & (0.67\%) & 0.3552 & (1.02\%) & 0.7774 & 0.2925 & 0.4965 & 1.148 & 0.3094 \\

\midrule
FCOS3D \cite{wang2021fcos3d} & R101 & - & - & 0.415 & - & 0.343 & - & 0.725 & 0.263 & 0.422 & 1.292 & 0.153 \\
PGD \cite{wang2021probabilistic} & R101 & - & - & 0.428 & - & 0.369 & - & 0.683 & 0.260 & 0.439 & 1.268 & 0.185 \\
DETR3D \cite{wang2021detrd} & R101 & - & - & 0.425 & - & 0.346 & - & 0.773 & 0.268 & 0.383 & 0.842 & 0.216 \\

\midrule
\rowcolor{gray95}
BEVFormer-base & R101 & - & - & 0.5056 & - & 0.4071 & - & 0.6843 & 0.278 & 0.3953 & 0.4241 & 0.1973 \\ 
\rowcolor{gray9}
BEVFormer-base-DE & R101 & 3.0 & 5.0 & \textbf{0.5069} & (0.26\%) & \textbf{0.4084} & (0.32\%) & 0.6854 & 0.2775 & 0.3911 & 0.4225 & 0.1969 \\ 

% \midrule
% \rowcolor{gray95}
% BEVFusion-CamOnly & Swin-T & - & - & 0.3954 & - & 0.3253 & - & 0.6992 & 0.2778 & 0.6135 & 0.8205 & 0.2612 \\ 
% \rowcolor{gray9}
% BEVFusion-CamOnly-DE & Swin-T & 0.3 & 5.0 & \textbf{0.4008} & (1.36\%) & \textbf{0.3287} & (1.05\%) & 0.6709 & 0.2691 & 0.5958 & 0.8440 & 0.2558 \\ 

\bottomrule
\end{tabular}
\label{tab:nuscenes-val}
% \vspace{-0.1in}
\end{table*}

\begin{table*}[!h]
\centering
\caption{\textbf{3D detection results with BEVFusion on nuScenes \texttt{val} set.} BevFusion-C stands for the \texttt{BevFusion-Camera only} variant, BevFusion-C-NS stands for \texttt{BevFusion-Camera Only No class balanced sampling}. The suffix \texttt{DE} signifies our method of applying \textit{Generalized Focal Loss} weights to each sample. SwinT in the table is refers to the Swin Transformer backbone. Best NDS and mAP scores are highlighted in \textbf{bold}.}
\setlength\tabcolsep{1.8pt}
\begin{tabular}{l c c c | c c | c c | c c c c c c }
\toprule
Method & Backbone & $\eta$ & $\gamma$ & NDS $\uparrow$ & $\Delta$ & mAP $\uparrow$ & $\Delta$  & mATE$\downarrow$     & mASE$\downarrow$     & mAOE$\downarrow$     & mAVE$\downarrow$    & mAAE$\downarrow$    \\

\midrule
\rowcolor{gray95}
BEVFusion-C & Swin-T & - & - & 0.3954 & - & 0.3253 & - & 0.6992 & 0.2778 & 0.6135 & 0.8205 & 0.2612 \\ 
\rowcolor{gray9}
BEVFusion-C-DE & Swin-T & 0.3 & 5.0 & \textbf{0.4008} & (1.36\%) & \textbf{0.3287} & (1.05\%) & 0.6709 & 0.2691 & 0.5958 & 0.8440 & 0.2558 \\ 

\midrule

\rowcolor{gray95}
BEVFusion-C-NC & Swin-T & - & - & 0.3382 & - & \textbf{0.3196} & - & 0.7630 & 0.2791 & 0.7762 & 1.2526 & 0.3970 \\
\rowcolor{gray9}
% BEVFormer-sml-DE & R101  & 0.5 & 5.0 & 0.3886 & 0.3509 & 0.7984 & 0.2932 & 0.4878 & 1.109 & 0.2892 \\
\rowcolor{gray9}
BEVFusion-C-NC-DE & Swin-T  & 0.3 & 5.0 & \textbf{0.3741} & (10.61\%) & 0.3114 & (-2.56\%) & 0.7760 & 0.2699 & 0.5314 & 0.9467 & 0.2918 \\
\rowcolor{gray9}

\bottomrule
\end{tabular}
\label{tab:nuscenes-val-bevfusion}
\vspace{-0.1in}
\end{table*}

% \begin{table*}[!h]
% \centering
% \caption{\textbf{3D detection results on nuScenes \texttt{val} set with/without class balancing.} BEVFusion-CamOnly stands for the \texttt{BEVFusion-Camera only} variant, and the suffix \texttt{DE} signifies our method of applying \textit{Generalized Focal Loss} weights to each sample. Swin-T in the table refers to the Swin Transformer. Best NDS and mAP scores are highlighted in \textbf{bold}.}
% \setlength\tabcolsep{2.2pt}
% \begin{tabular}{l c c c | c c | c c | c c c c c c }
% \toprule
% Method & Backbone & $\eta$ & $\gamma$ & NDS$\uparrow$ & \Delta & mAP$\uparrow$ & \Delta  & mATE$\downarrow$     & mASE$\downarrow$     & mAOE$\downarrow$     & mAVE$\downarrow$    & mAAE$\downarrow$    \\
% \midrule

% \rowcolor{gray95}
% BEVFusion-CamOnly & Swin-T & - & - & 0.3382 & - & \textbf{0.3196} & - & 0.7630 & 0.2791 & 0.7762 & 1.2526 & 0.3970 \\
% \rowcolor{gray9}
% % BEVFormer-sml-DE & R101  & 0.5 & 5.0 & 0.3886 & 0.3509 & 0.7984 & 0.2932 & 0.4878 & 1.109 & 0.2892 \\
% \rowcolor{gray9}
% BEVFusion-CamOnly-DE & Swin-T  & 0.3 & 5.0 & \textbf{0.3741} & (10.61\%) & 0.3114 & (-2.56\%) & 0.7760 & 0.2699 & 0.5314 & 0.9467 & 0.2918 \\
% \rowcolor{gray9}

% \bottomrule
% \end{tabular}
% \label{tab:nuscenes-val-futurework}
% \end{table*}

\raggedbottom
BEVFusion camera only model has been trained from scratch as the baseline model, which uses Swin Transformer pretrained on nuImages as the backbone network. With the official code-base \footnote{BEVfusion official implementation: \url{https://github.com/mit-han-lab/bevfusion}}, we have been able to obtain similar baseline results ($0.3954$ NDS), and further improve them with \textit{Generalized Focal Loss}. Detailed results are in Table \ref{tab:nuscenes-val-bevfusion}. 

For all experiments with BEVFormer, $16$ NVIDIA A100 GPUs were used, and the models were trained with a learning rate of $2 \times 10^{-4}$ along with \textit{CosineAnnealing} scheduler for a total of $24$ epochs.  For BEVFusion, all experiments used 8 NVIDIA A100 GPUS trained for 20 epochs.

\textbf{Dataset Bias Quantification:}
For Waymo Dataset Analysis, our experiments have been conducted on the front camera subset($200,000$) of the Waymo training set split(1 million images). Using \texttt{resnet101} for feature extraction, followed by TSNE and DBSCAN to compute clusters for lower dimensional space. As indicated in Fig.\ref{fig:cluster-sampled-likelihood}-(d), with the visualization on most likely clusters of dataset samples, several semantically meaningful clusters have formed, namely cluster $92$ of city crosswalk, cluster $62$ of the residential driving scene of palm trees, cluster $39$ of crowded driving scenes, and cluster 32 of construction vehicles and traffic cones. Similarly, the BDD100K dataset contains mostly two scenarios - daylight and nighttime( Fig.\ref{fig:cluster-sampled-likelihood} -(c) ). 

% In TuSimple lane detection dataset, all \textit{train}, \textit{test}, and \textit{val} splits were considered for dataset bias quantification. After feature clustering, it can be observed that almost all the samples fall under one cluster. This can be due to the dataset being inherently biased towards a particular environment (highway) and conditions (daylight and normal weather). In addition, the image frames are sequential in nature which further makes it hard to find semantically meaningful clusters. More discussion about this dataset is in the supplementary section.
%Semantic clustering also reveals two major clusters for the said frames and a few small clusters representing outlier scenarios of high reflections from brake lights, sun flare, snowy nights, etc.
% \vspace{1.5mm}

%% Further Discussions
\vspace{-0.1in}
\section{Further Discussion}

It is worth noting that the proposed generalized focal loss function has been tested end-to-end on two autonomous driving datasets and in the context of 3D object detection. Semantics-based dataset distribution has been analyzed and bias has been quantified for $2$ additional datasets, and the results demonstrate that those datasets could also benefit from our loss function. It is possible that this loss function may also be effective for other tasks, but this has not yet been explored. Parameters used in this paper for \texttt{t-SNE} and \texttt{DBSCAN} were chosen analytically by visualizing the cluster and samples from each of them.

\vspace{-0.1in}

%% Conclusion

\section{Conclusion}
% Dataset bias is a huge issue in machine learning and is not limited to the class imbalance problem. The appearance of image samples could also induce bias in the training process depending on the likelihood of such samples appearing in the dataset. This paper proposes a novel method to address this issue in machine learning, particularly in computer vision. First, likelihood is computed for each sample within the dataset using image semantics. Then the proposed loss function called, \textit{Generalized Focal Loss}, is used to weigh loss for each sample in a way that enforces equity within the dataset during training. This method is simple, intuitive, and proves to be exceedingly effective for under-represented samples within datasets. We hope that our research provides a useful add-on to future researchers towards minimizing implicit dataset bias.

While prior work tackles explicit class imbalance, this paper addresses the overlooked issue of implicit dataset bias caused by non-uniform sample likelihoods. By modeling raw image data instead of just categorical labels, the proposed likelihood computation and \textit{Generalized Focal Loss} reweight training in a ground-truth agnostic manner. Experiments demonstrate improved rare and out-of-distribution sample modeling over state-of-the-art techniques relying solely on class frequencies. The consistent gains highlight the benefits of understanding dataset heterogeneity beyond class labels alone. This likelihood-based approach provides a useful complement to existing class-aware balancing methods. By countering biases in the raw visual distribution itself, it opens promising new directions for unsupervised analysis and optimization. The principles explored in this work can enable future research on minimizing implicit dataset imbalances.

\clearpage

{\small
\bibliographystyle{ieee_fullname}
\bibliography{references}

\begin{thebibliography}{10}\itemsep=-1pt

\bibitem{8658633}
Björn Barz and Joachim Denzler.
\newblock Hierarchy-based image embeddings for semantic image retrieval.
\newblock In {\em 2019 IEEE Winter Conference on Applications of Computer
  Vision (WACV)}, pages 638--647, 2019.

\bibitem{10.1145/2907070}
Paula Branco, Lu\'{\i}s Torgo, and Rita~P. Ribeiro.
\newblock A survey of predictive modeling on imbalanced domains.
\newblock {\em ACM Comput. Surv.}, 49(2), aug 2016.

\bibitem{brazil2019m3drpn}
Garrick Brazil and Xiaoming Liu.
\newblock M3d-rpn: Monocular 3d region proposal network for object detection.
\newblock In {\em Proceedings of the IEEE International Conference on Computer
  Vision}, Seoul, South Korea, 2019.

\bibitem{importance-weighing}
Jonathon Byrd and Zachary~C. Lipton.
\newblock Weighted risk minimization {\&} deep learning.
\newblock {\em CoRR}, abs/1812.03372, 2018.

\bibitem{nuscenes}
Holger Caesar, Varun Bankiti, Alex~H. Lang, Sourabh Vora, Venice~Erin Liong,
  Qiang Xu, Anush Krishnan, Yu Pan, Giancarlo Baldan, and Oscar Beijbom.
\newblock nuscenes: A multimodal dataset for autonomous driving.
\newblock In {\em CVPR}, 2020.

\bibitem{campello2013density}
Ricardo~JGB Campello, Davoud Moulavi, and J{\"o}rg Sander.
\newblock Density-based clustering based on hierarchical density estimates.
\newblock In {\em Advances in Knowledge Discovery and Data Mining: 17th
  Pacific-Asia Conference, PAKDD 2013, Gold Coast, Australia, April 14-17,
  2013, Proceedings, Part II 17}, pages 160--172. Springer, 2013.

\bibitem{caron2021emerging}
Mathilde Caron, Hugo Touvron, Ishan Misra, Herv\'e J\'egou, Julien Mairal,
  Piotr Bojanowski, and Armand Joulin.
\newblock Emerging properties in self-supervised vision transformers.
\newblock In {\em Proceedings of the International Conference on Computer
  Vision (ICCV)}, 2021.

\bibitem{NIPS2015_6da37dd3}
Xiaozhi Chen, Kaustav Kundu, Yukun Zhu, Andrew~G Berneshawi, Huimin Ma, Sanja
  Fidler, and Raquel Urtasun.
\newblock 3d object proposals for accurate object class detection.
\newblock In C. Cortes, N. Lawrence, D. Lee, M. Sugiyama, and R. Garnett,
  editors, {\em Advances in Neural Information Processing Systems}, volume~28.
  Curran Associates, Inc., 2015.

\bibitem{cui2019classbalancedloss}
Yin Cui, Menglin Jia, Tsung-Yi Lin, Yang Song, and Serge Belongie.
\newblock Class-balanced loss based on effective number of samples.
\newblock In {\em CVPR}, 2019.

\bibitem{dbscan}
Martin Ester, Hans-Peter Kriegel, J\"{o}rg Sander, and Xiaowei Xu.
\newblock A density-based algorithm for discovering clusters in large spatial
  databases with noise.
\newblock In {\em Proceedings of the Second International Conference on
  Knowledge Discovery and Data Mining}, KDD'96, page 226–231. AAAI Press,
  1996.

\bibitem{5539906}
Pedro~F. Felzenszwalb, Ross~B. Girshick, and David McAllester.
\newblock Cascade object detection with deformable part models.
\newblock In {\em 2010 IEEE Computer Society Conference on Computer Vision and
  Pattern Recognition}, pages 2241--2248, 2010.

\bibitem{9324926}
K.~Ruwani~M. Fernando and Chris~P. Tsokos.
\newblock Dynamically weighted balanced loss: Class imbalanced learning and
  confidence calibration of deep neural networks.
\newblock {\em IEEE Transactions on Neural Networks and Learning Systems},
  33(7):2940--2951, 2022.

\bibitem{9093421}
Siddhartha Gairola, Rajvi Shah, and P.J. Narayanan.
\newblock Unsupervised image style embeddings for retrieval and recognition
  tasks.
\newblock In {\em 2020 IEEE Winter Conference on Applications of Computer
  Vision (WACV)}, pages 3270--3278, 2020.

\bibitem{Geiger2012CVPR}
Andreas Geiger, Philip Lenz, and Raquel Urtasun.
\newblock Are we ready for autonomous driving? the kitti vision benchmark
  suite.
\newblock In {\em Conference on Computer Vision and Pattern Recognition
  (CVPR)}, 2012.

\bibitem{5128907}
Haibo He and Edwardo~A. Garcia.
\newblock Learning from imbalanced data.
\newblock {\em IEEE Transactions on Knowledge and Data Engineering},
  21(9):1263--1284, 2009.

\bibitem{resnet}
Kaiming He, Xiangyu Zhang, Shaoqing Ren, and Jian Sun.
\newblock Deep residual learning for image recognition.
\newblock {\em CoRR}, abs/1512.03385, 2015.

\bibitem{7780459}
Kaiming He, Xiangyu Zhang, Shaoqing Ren, and Jian Sun.
\newblock Deep residual learning for image recognition.
\newblock In {\em 2016 IEEE Conference on Computer Vision and Pattern
  Recognition (CVPR)}, pages 770--778, 2016.

\bibitem{huang2021bevdet}
Junjie Huang, Guan Huang, Zheng Zhu, Ye Yun, and Dalong Du.
\newblock Bevdet: High-performance multi-camera 3d object detection in
  bird-eye-view.
\newblock {\em arXiv preprint arXiv:2112.11790}, 2021.

\bibitem{9857264}
Nikita Jaipuria, Katherine Stevo, Xianling Zhang, Meghana~L. Gaopande,
  Ian~Calle Garcia, Jinesh Jain, and Vidya~N. Murali.
\newblock deeppic: Deep perceptual image clustering for identifying bias in
  vision datasets.
\newblock In {\em 2022 IEEE/CVF Conference on Computer Vision and Pattern
  Recognition Workshops (CVPRW)}, pages 4792--4801, 2022.

\bibitem{Jaipuria_2020_CVPR_Workshops}
Nikita Jaipuria, Xianling Zhang, Rohan Bhasin, Mayar Arafa, Punarjay
  Chakravarty, Shubham Shrivastava, Sagar Manglani, and Vidya~N. Murali.
\newblock Deflating dataset bias using synthetic data augmentation.
\newblock In {\em Proceedings of the IEEE/CVF Conference on Computer Vision and
  Pattern Recognition (CVPR) Workshops}, June 2020.

\bibitem{Jolliffe:1986}
I.T. Jolliffe.
\newblock {\em Principal Component Analysis}.
\newblock Springer Verlag, 1986.

\bibitem{krishnan2023lane}
Akshay Krishnan, Amit Raj, Xianling Zhang, Alexandra Carlson, Nathan Tseng,
  Sandhya Sridhar, Nikita Jaipuria, and James Hays.
\newblock Lane: Lighting-aware neural fields for compositional scene synthesis,
  2023.

\bibitem{lee2019centermask}
Youngwan Lee and Jongyoul Park.
\newblock Centermask: Real-time anchor-free instance segmentation.
\newblock 2020.

\bibitem{li2022deepfusion}
Yingwei Li, Adams~Wei Yu, Tianjian Meng, Ben Caine, Jiquan Ngiam, Daiyi Peng,
  Junyang Shen, Yifeng Lu, Denny Zhou, Quoc~V Le, et~al.
\newblock Deepfusion: Lidar-camera deep fusion for multi-modal 3d object
  detection.
\newblock In {\em Proceedings of the IEEE/CVF Conference on Computer Vision and
  Pattern Recognition}, pages 17182--17191, 2022.

\bibitem{li2022bevformer}
Zhiqi Li, Wenhai Wang, Hongyang Li, Enze Xie, Chonghao Sima, Tong Lu, Qiao Yu,
  and Jifeng Dai.
\newblock Bevformer: Learning bird's-eye-view representation from multi-camera
  images via spatiotemporal transformers, 2022.

\bibitem{focal-loss}
Tsung{-}Yi Lin, Priya Goyal, Ross~B. Girshick, Kaiming He, and Piotr
  Doll{\'{a}}r.
\newblock Focal loss for dense object detection.
\newblock {\em CoRR}, abs/1708.02002, 2017.

\bibitem{linderman2019clustering}
George~C Linderman and Stefan Steinerberger.
\newblock Clustering with t-sne, provably.
\newblock {\em SIAM Journal on Mathematics of Data Science}, 1(2):313--332,
  2019.

\bibitem{liu2022bevfusion}
Zhijian Liu, Haotian Tang, Alexander Amini, Xingyu Yang, Huizi Mao, Daniela
  Rus, and Song Han.
\newblock Bevfusion: Multi-task multi-sensor fusion with unified bird's-eye
  view representation.
\newblock {\em arXiv}, 2022.

\bibitem{hdbscan}
Leland McInnes, John Healy, and Steve Astels.
\newblock hdbscan: Hierarchical density based clustering.
\newblock {\em Journal of Open Source Software}, 2(11):205, 2017.

\bibitem{pang2020clocs}
Su Pang, Daniel Morris, and Hayder Radha.
\newblock Clocs: Camera-lidar object candidates fusion for 3d object detection.
\newblock 2020.

\bibitem{park2021dd3d}
Dennis Park, Rares Ambrus, Vitor Guizilini, Jie Li, and Adrien Gaidon.
\newblock Is pseudo-lidar needed for monocular 3d object detection?
\newblock In {\em IEEE/CVF International Conference on Computer Vision (ICCV)},
  2021.

\bibitem{radford2021learning}
Alec Radford, Jong~Wook Kim, Chris Hallacy, Aditya Ramesh, Gabriel Goh,
  Sandhini Agarwal, Girish Sastry, Amanda Askell, Pamela Mishkin, Jack Clark,
  et~al.
\newblock Learning transferable visual models from natural language
  supervision.
\newblock In {\em International Conference on Machine Learning}, pages
  8748--8763. PMLR, 2021.

\bibitem{clip}
Alec Radford, Jong~Wook Kim, Chris Hallacy, Aditya Ramesh, Gabriel Goh,
  Sandhini Agarwal, Girish Sastry, Amanda Askell, Pamela Mishkin, Jack Clark,
  Gretchen Krueger, and Ilya Sutskever.
\newblock Learning transferable visual models from natural language
  supervision.
\newblock {\em CoRR}, abs/2103.00020, 2021.

\bibitem{NIPS2015_14bfa6bb}
Shaoqing Ren, Kaiming He, Ross Girshick, and Jian Sun.
\newblock Faster r-cnn: Towards real-time object detection with region proposal
  networks.
\newblock In C. Cortes, N. Lawrence, D. Lee, M. Sugiyama, and R. Garnett,
  editors, {\em Advances in Neural Information Processing Systems}, volume~28.
  Curran Associates, Inc., 2015.

\bibitem{journals/corr/ShrivastavaGG16}
Abhinav Shrivastava, Abhinav Gupta, and Ross~B. Girshick.
\newblock Training region-based object detectors with online hard example
  mining.
\newblock {\em CoRR}, abs/1604.03540, 2016.

\bibitem{shrivastava2021vr3dense}
Shubham Shrivastava.
\newblock Vr3dense: Voxel representation learning for 3d object detection and
  monocular dense depth reconstruction.
\newblock {\em arXiv preprint arXiv:2104.05932}, 2021.

\bibitem{cubifae3d}
Shubham Shrivastava and Punarjay Chakravarty.
\newblock Cubifae-3d: Monocular camera space cubification for auto-encoder
  based 3d object detection, 2020.

\bibitem{simonelli2019disentangling}
Andrea Simonelli, Samuel~Rota Bulo, Lorenzo Porzi, Manuel L{\'o}pez-Antequera,
  and Peter Kontschieder.
\newblock Disentangling monocular 3d object detection.
\newblock In {\em Proceedings of the IEEE/CVF International Conference on
  Computer Vision}, pages 1991--1999, 2019.

\bibitem{Simonelli2019DisentanglingM3}
Andrea Simonelli, Samuel~Rota Bul{\`o}, Lorenzo Porzi, Manuel
  L{\'o}pez-Antequera, and Peter Kontschieder.
\newblock Disentangling monocular 3d object detection.
\newblock {\em 2019 IEEE/CVF International Conference on Computer Vision
  (ICCV)}, pages 1991--1999, 2019.

\bibitem{9200697}
Andrea Simonelli, Samuel~Rota Bulò, Lorenzo Porzi, Manuel~López Antequera,
  and Peter Kontschieder.
\newblock Disentangling monocular 3d object detection: From single to
  multi-class recognition.
\newblock {\em IEEE Transactions on Pattern Analysis and Machine Intelligence},
  44(3):1219--1231, 2022.

\bibitem{Sun_2020_CVPR}
Pei Sun, Henrik Kretzschmar, Xerxes Dotiwalla, Aurelien Chouard, Vijaysai
  Patnaik, Paul Tsui, James Guo, Yin Zhou, Yuning Chai, Benjamin Caine, Vijay
  Vasudevan, Wei Han, Jiquan Ngiam, Hang Zhao, Aleksei Timofeev, Scott
  Ettinger, Maxim Krivokon, Amy Gao, Aditya Joshi, Yu Zhang, Jonathon Shlens,
  Zhifeng Chen, and Dragomir Anguelov.
\newblock Scalability in perception for autonomous driving: Waymo open dataset.
\newblock In {\em Proceedings of the IEEE/CVF Conference on Computer Vision and
  Pattern Recognition (CVPR)}, June 2020.

\bibitem{10.5555/929901}
Kah~Kay Sung and Tomaso~A. Poggio.
\newblock {\em Learning and Example Selection for Object and Pattern
  Detection}.
\newblock PhD thesis, USA, 1996.
\newblock AAI0800657.

\bibitem{tsne}
Laurens van~der Maaten and Geoffrey Hinton.
\newblock Visualizing data using t-sne.
\newblock {\em Journal of Machine Learning Research}, 9(86):2579--2605, 2008.

\bibitem{990517}
P. Viola and M. Jones.
\newblock Rapid object detection using a boosted cascade of simple features.
\newblock In {\em Proceedings of the 2001 IEEE Computer Society Conference on
  Computer Vision and Pattern Recognition. CVPR 2001}, volume~1, pages I--I,
  2001.

\bibitem{9081913}
Lin Wang, Chaoli Wang, Zhanquan Sun, Shuqun Cheng, and Lei Guo.
\newblock Class balanced loss for image classification.
\newblock {\em IEEE Access}, 8:81142--81153, 2020.

\bibitem{wang2021fcos3d}
Tai Wang, Xinge Zhu, Jiangmiao Pang, and Dahua Lin.
\newblock Fcos3d: Fully convolutional one-stage monocular 3d object detection.
\newblock In {\em IEEE/CVF Conference on International Conference on Computer
  Vision Workshops}, 2021.

\bibitem{wang2021probabilistic}
Tai Wang, Xinge Zhu, Jiangmiao Pang, and Dahua Lin.
\newblock Probabilistic and geometric depth: Detecting objects in perspective.
\newblock In {\em Conference on Robot Learning}, 2021.

\bibitem{wang2021detrd}
Yue Wang, Vitor~Campagnolo Guizilini, Tianyuan Zhang, Yilun Wang, Hang Zhao,
  and Justin Solomon.
\newblock {DETR}3d: 3d object detection from multi-view images via 3d-to-2d
  queries.
\newblock In {\em 5th Annual Conference on Robot Learning}, 2021.

\bibitem{yu2020bdd100k}
Fisher Yu, Haofeng Chen, Xin Wang, Wenqi Xian, Yingying Chen, Fangchen Liu,
  Vashisht Madhavan, and Trevor Darrell.
\newblock Bdd100k: A diverse driving dataset for heterogeneous multitask
  learning.
\newblock In {\em Proceedings of the IEEE/CVF conference on computer vision and
  pattern recognition}, pages 2636--2645, 2020.

\bibitem{Zhang_2022_CVPR}
Xianling Zhang, Nathan Tseng, Ameerah Syed, Rohan Bhasin, and Nikita Jaipuria.
\newblock Simbar: Single image-based scene relighting for effective data
  augmentation for automated driving vision tasks.
\newblock In {\em Proceedings of the IEEE/CVF Conference on Computer Vision and
  Pattern Recognition (CVPR)}, pages 3718--3728, June 2022.

\bibitem{zhang2022beverse}
Yunpeng Zhang, Zheng Zhu, Wenzhao Zheng, Junjie Huang, Guan Huang, Jie Zhou,
  and Jiwen Lu.
\newblock Beverse: Unified perception and prediction in birds-eye-view for
  vision-centric autonomous driving.
\newblock {\em arXiv preprint arXiv:2205.09743}, 2022.

\bibitem{zhu2019classbalanced}
Benjin Zhu, Zhengkai Jiang, Xiangxin Zhou, Zeming Li, and Gang Yu.
\newblock Class-balanced grouping and sampling for point cloud 3d object
  detection, 2019.

\end{thebibliography}
}
%%%%%%%%%%%%%%%%%%%%%%%%%%%%%%%%%%%%%
% APPENDIX
%%%%%%%%%%%%%%%%%%%%%%%%%%%%%%%%%%%%%
\newpage
\appendix
\onecolumn
\renewcommand{\thefigure}{S\arabic{figure}}
\setcounter{figure}{0}

\begin{center}
\vspace{2in}
\Large\bfseries
Supplementary Materials \\DatasetEquity: Are All Samples Created Equal? \\In The Quest For Equity Within Datasets
\\[3\baselineskip]
\end{center}

\section{Dataset cluster visualization}

\begin{figure*}[h]
\begin{center}

\includegraphics[width=0.7\linewidth]{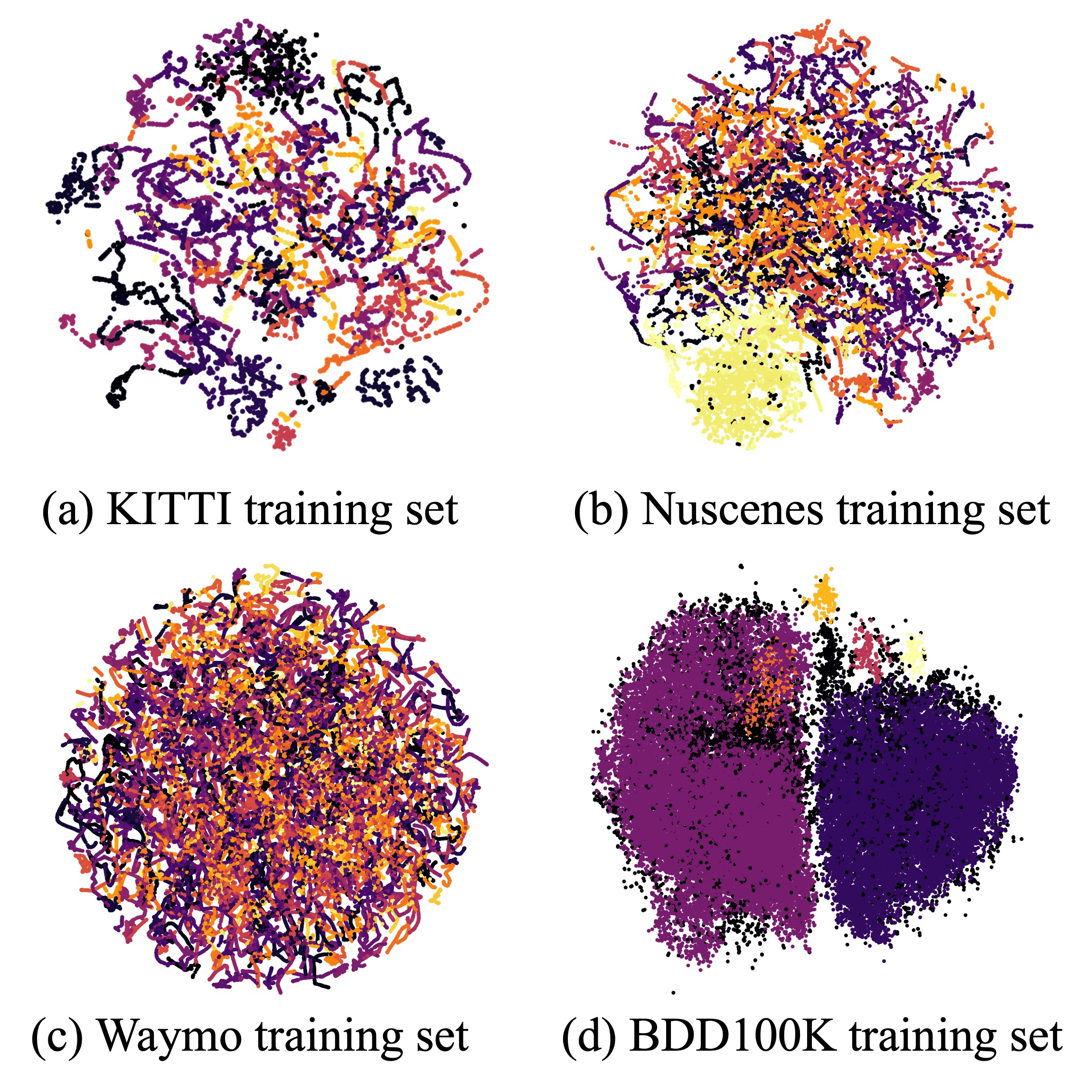}
\end{center}
   \caption{NuScenes, KITTI, Waymo, and BDD100k dataset samples projected onto a 3-dimensional t-SNE space and then clustered using DBSCAN (only first 2-dimensions are visualized). Each color represents a unique cluster ID. Only a single front camera image was used to compute these clusters.
}
\label{fig:cluster_visualization}
\end{figure*}

Camera images from various datasets are first passed through a feature extractor, followed by a low-dimensional feature projection (\texttt{t-SNE}). These low-dimensional (3D) features are then clustered together to reveal frames with similar semantics and perceptual details. The relative size of each of these clusters is then used as a proxy for quantifying the likelihood of occurrence for each sample within those clusters. Figure \ref{fig:cluster_visualization} visualizes these clusters in 2D for the four datasets analyzed in this work.
\vspace{0.3in}

\section{Dataset samples visualization}

A few examples of samples from various clusters and their scaled likelihoods are shown in sections \ref{sec:kitti-supplementary}, \ref{sec:nuscenes-supplementary}, \ref{sec:waymo-supplementary}, and \ref{sec:bdd100k-supplementary}. These clusters were computed by first projecting $2048$-dimensional image embeddings onto a lower $3$-dimensional space using t-SNE, and then applying the DBSCAN algorithm. The likelihood of each sample is then computed as $\mathcal{L}_{s^{(i)}} = \frac{|C_i|}{\max_i (|C_i|)}$, where $C_i$ refers to the $i^{th}$ cluster, $|C_i|$ is the number of samples in the $i^{th}$ cluster, and $\mathcal{L}_{s^{(i)}}$ is the scaled likelihood of samples corresponding to the $i^{th}$ cluster. Corresponding to each row in the visualized samples below, we also provide the measure for corresponding likelihood values. All images in a certain row are taken from the same cluster.

\subsection{KITTI training dataset}\label{sec:kitti-supplementary}

\begin{figure*}[!h]  
  \centering
  \subfloat[$\mathcal{L}_{s^{(i)}}=0.0192$]
  {\includegraphics[width=\linewidth]{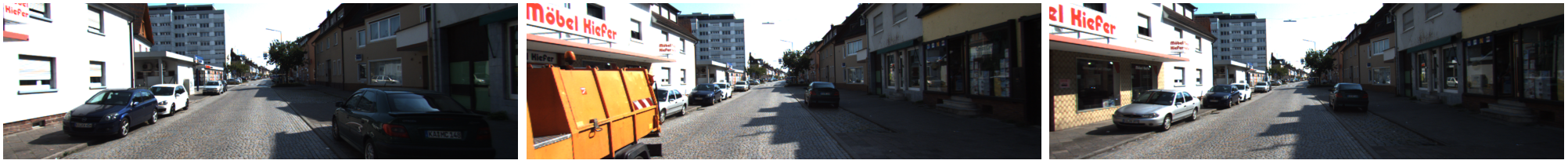}}

  \centering
  \subfloat[$\mathcal{L}_{s^{(i)}}=0.0240$]
  {\includegraphics[width=\linewidth]{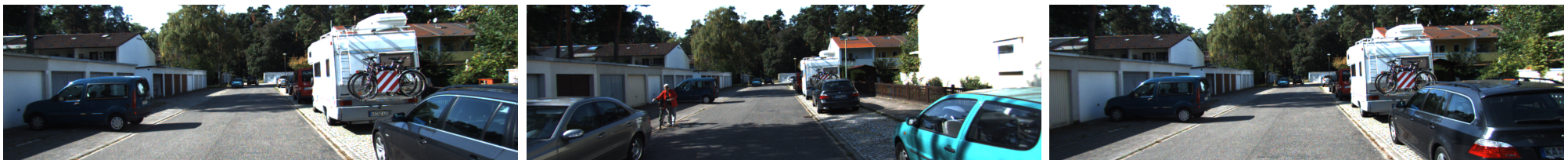}}

  \centering
  \subfloat[$\mathcal{L}_{s^{(i)}}=0.4855$]
  {\includegraphics[width=\linewidth]{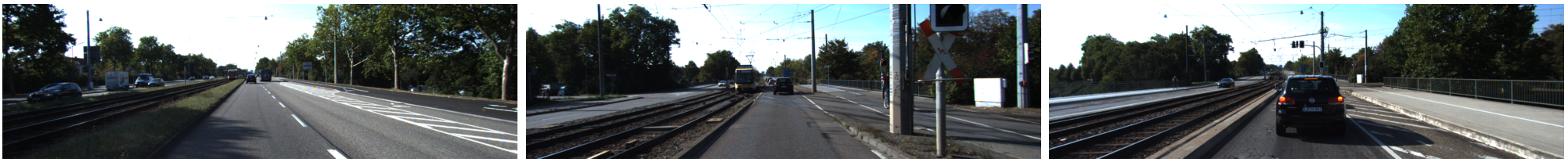}}
  
  \centering
  \subfloat[$\mathcal{L}_{s^{(i)}}=0.6178$]
  {\includegraphics[width=\linewidth]{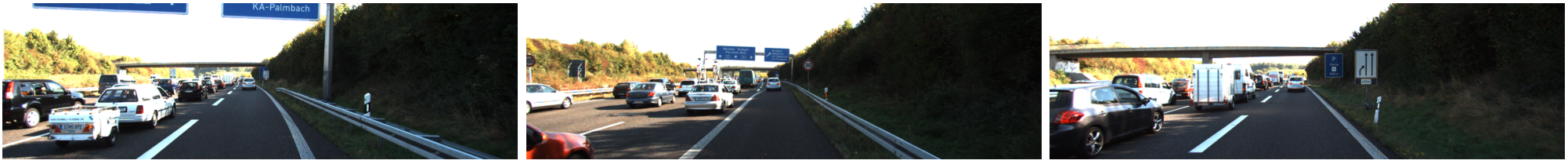}}

  \centering
  \subfloat[$\mathcal{L}_{s^{(i)}}=1.0$]
  {\includegraphics[width=\linewidth]{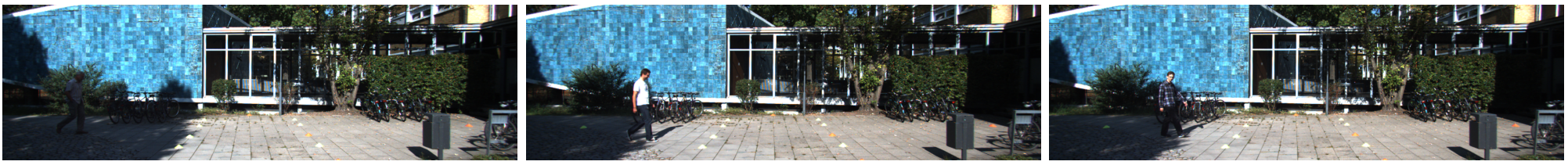}}

  \caption{Samples from various clusters in KITTI \textit{training} dataset, and their likelihoods relative to the largest cluster within the dataset. Samples shown in the last row come from a cluster of size $416$, and is the largest among all clusters.}
  \label{fig:kitti-samples-w-likelihood}
\end{figure*}

\clearpage
\section{nuScenes training dataset}\label{sec:nuscenes-supplementary}

\begin{figure*}[!h]  
  \centering
  \subfloat[$\mathcal{L}_{s^{(i)}}=0.0034$]
  {\includegraphics[width=\linewidth]{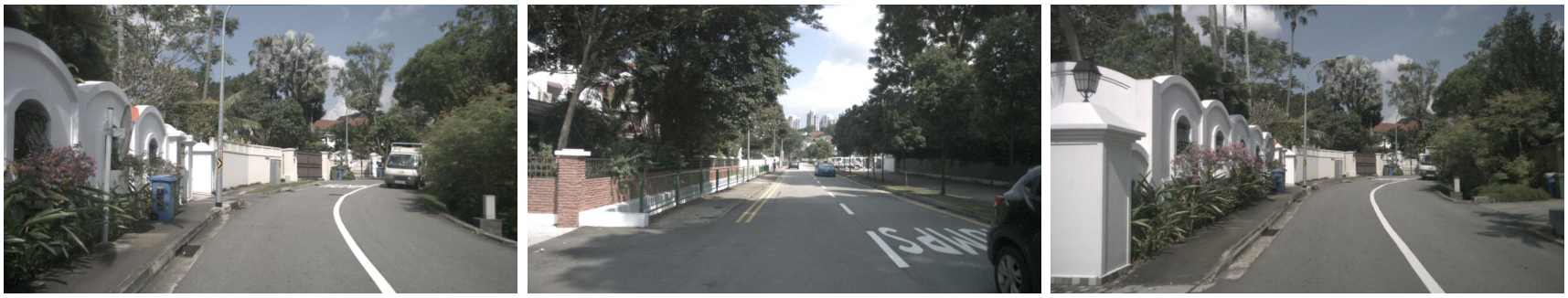}}

  \centering
  \subfloat[$\mathcal{L}_{s^{(i)}}=0.0034$]
  {\includegraphics[width=\linewidth]{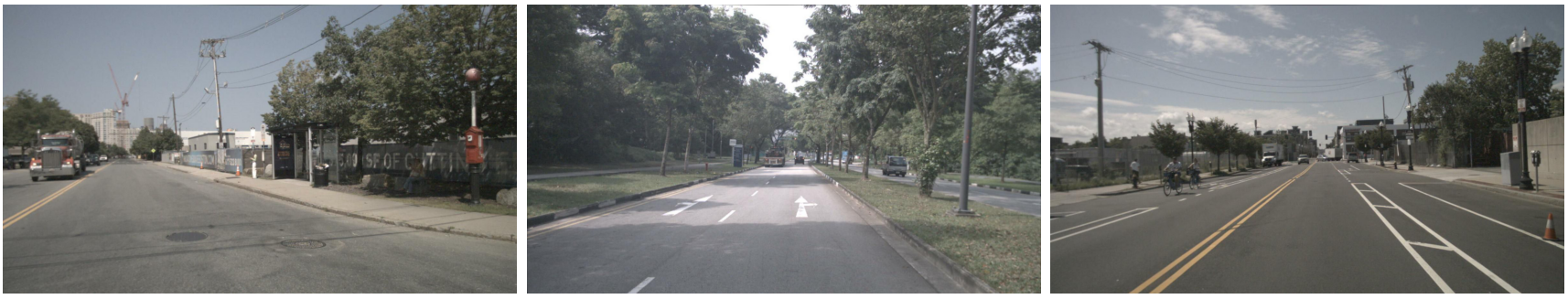}}

  \centering
  \subfloat[$\mathcal{L}_{s^{(i)}}=0.0038$]
  {\includegraphics[width=\linewidth]{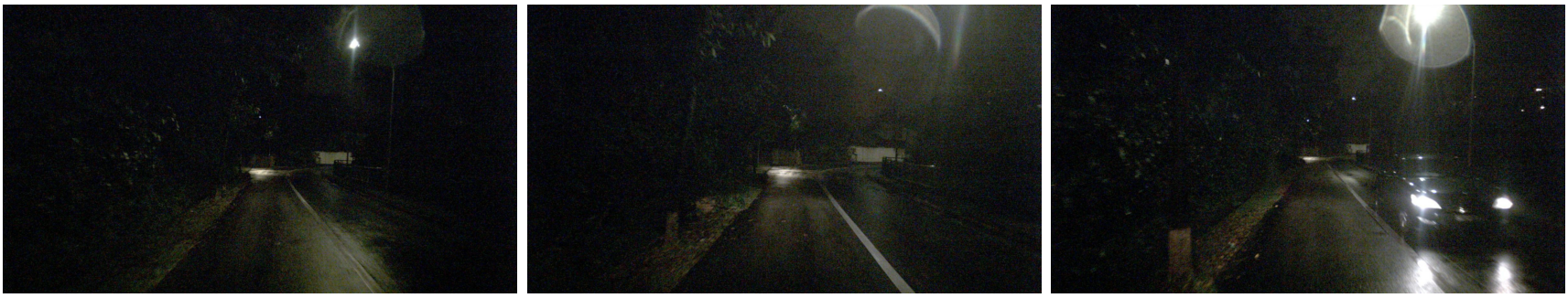}}
  
  \centering
  \subfloat[$\mathcal{L}_{s^{(i)}}=0.3592$]
  {\includegraphics[width=\linewidth]{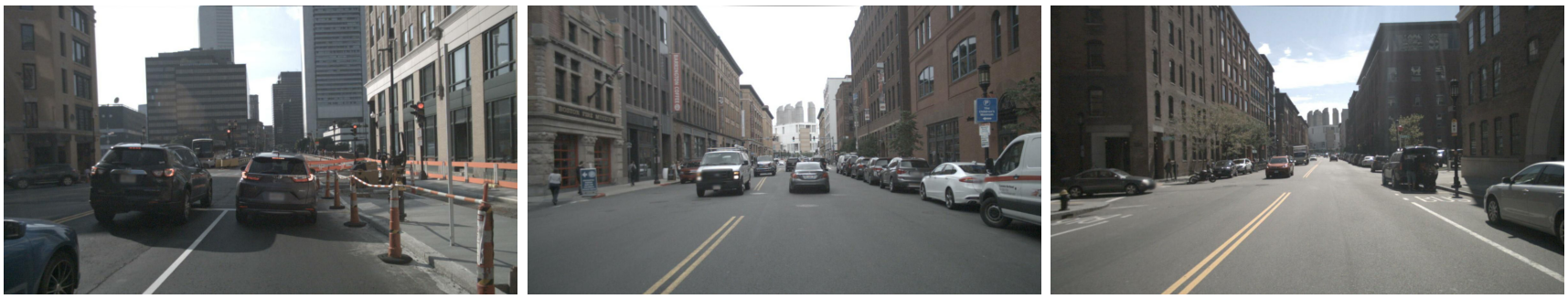}}

  \centering
  \subfloat[$\mathcal{L}_{s^{(i)}}=1.0$]
  {\includegraphics[width=\linewidth]{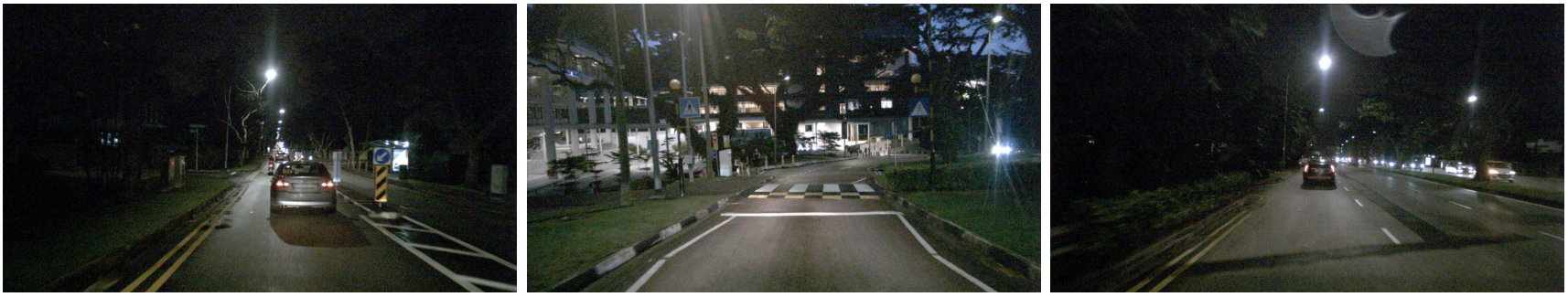}}

  \caption{Samples from various clusters in nuScenes \textit{training} dataset, and their likelihoods relative to the largest cluster within the dataset. Samples shown in the last row come from a cluster of size $2344$, and is the largest among all clusters.}
  \label{fig:kitti-samples-w-likelihood}
\end{figure*}

\newpage
\section{Waymo Open dataset}\label{sec:waymo-supplementary}
\begin{figure*}[!h]  
  \centering
  \subfloat[$\mathcal{L}_{s^{(i)}}=0.2487$]
  {\includegraphics[width=0.8\linewidth]{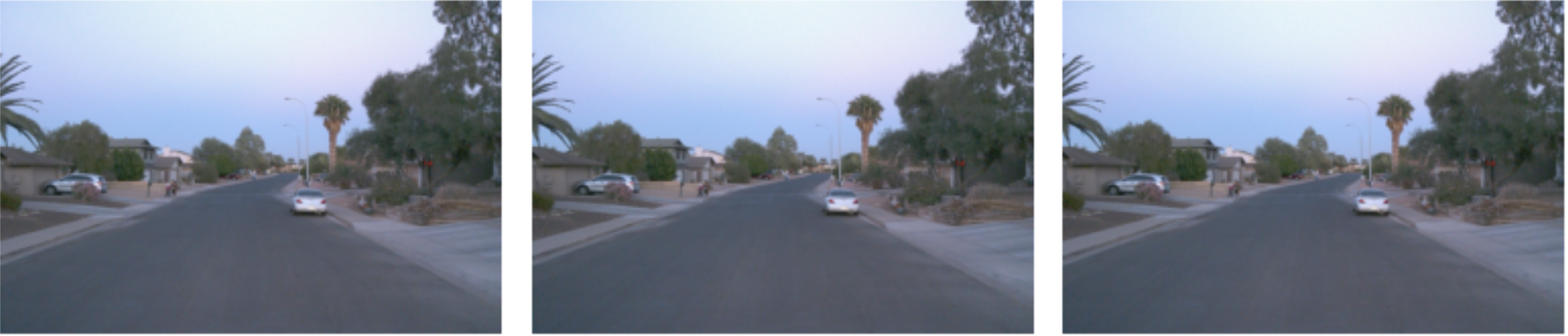}}
  
  \centering
  \subfloat[$\mathcal{L}_{s^{(i)}}=0.3178$]
  {\includegraphics[width=0.8\linewidth]{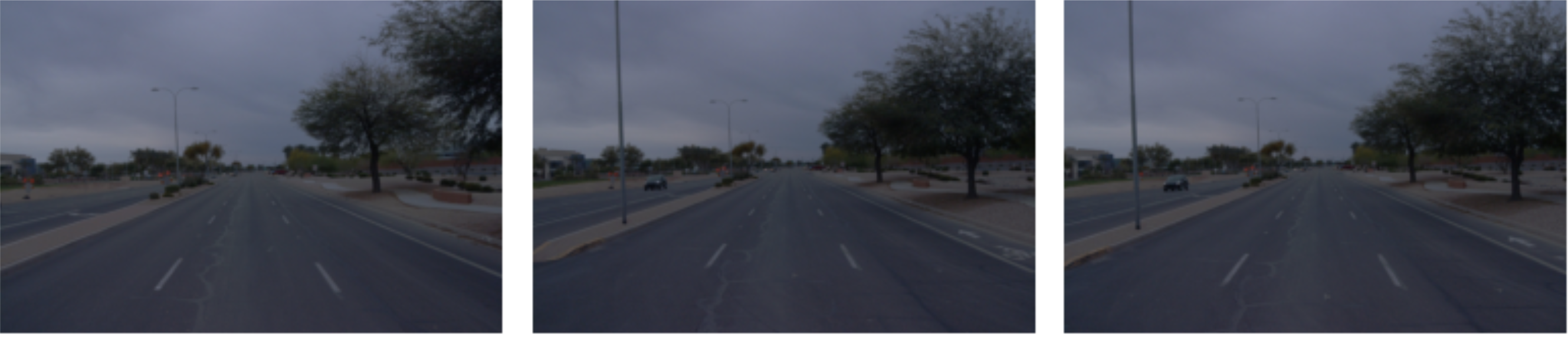}}

    \centering
  \subfloat[$\mathcal{L}_{s^{(i)}}=0.7123$]
  {\includegraphics[width=0.8\linewidth]{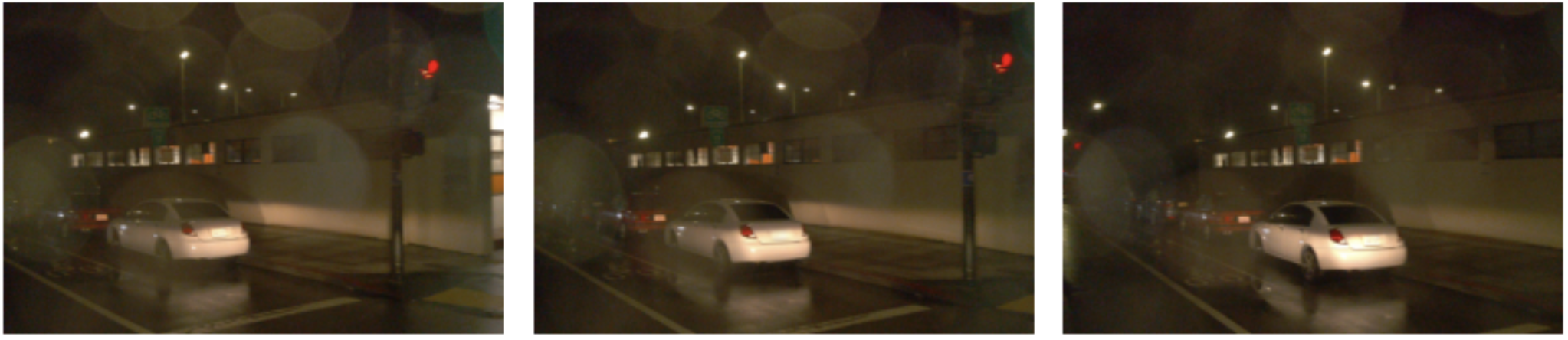}}

  \centering
  \subfloat[$\mathcal{L}_{s^{(i)}}=0.7462$]
  {\includegraphics[width=0.8\linewidth]{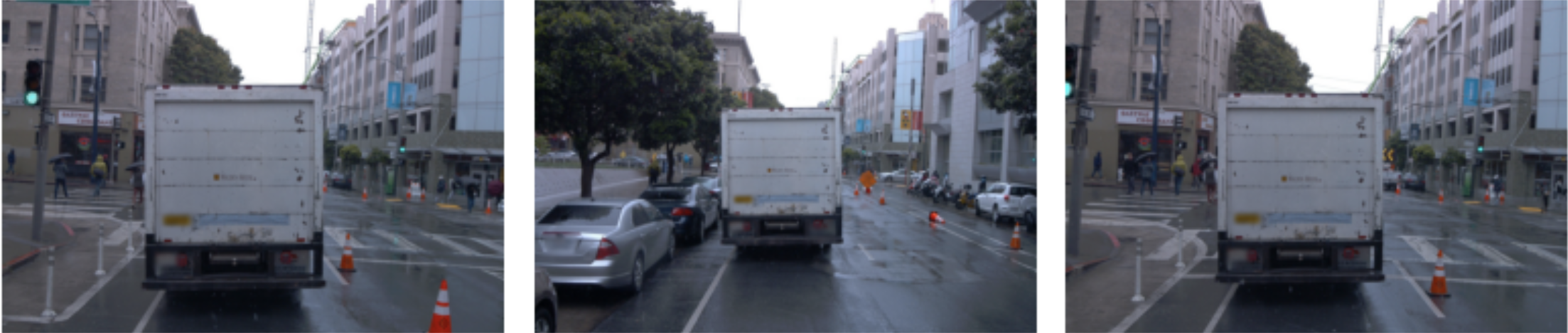}}
  
  \centering
  \subfloat[$\mathcal{L}_{s^{(i)}}=1.0$]
  {\includegraphics[width=0.8\linewidth]{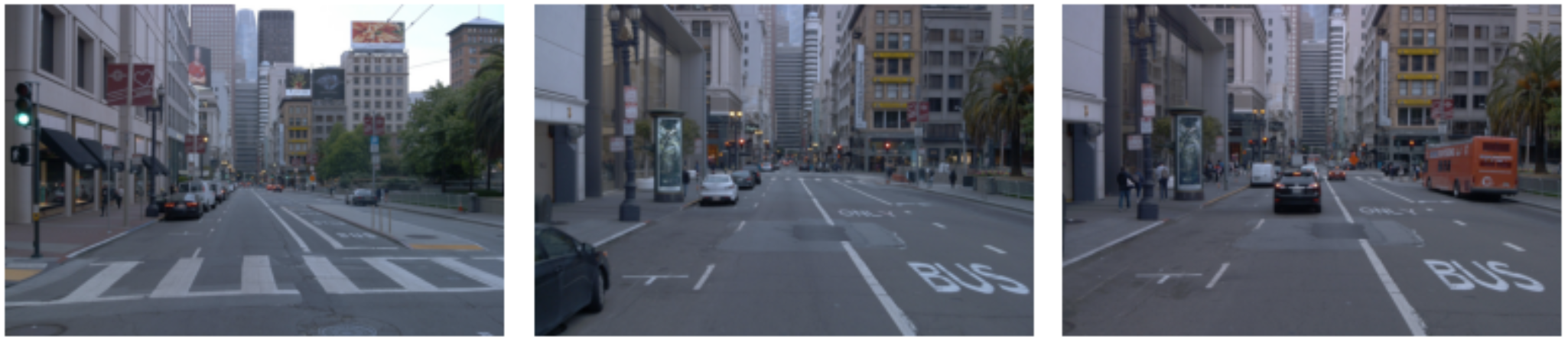}}
  
  \caption{Samples from various clusters in Waymo \textit{training} dataset, and their likelihoods relative to the largest cluster within the dataset. Totally, $1408$ clusters formed within the $200000$ training samples from the front camera. The rarest clusters contain (a) (b) and 
 (c) of only 10 samples each in the empty residential driving scene, after-sunset scene, and high rainy scene respectively. Samples shown in the last row come from cluster $87$ of size $796$, and is the largest among all clusters. The rest top largest clusters share similar cluster sizes of $500$ to $700$ samples from urban city driving scenes, which indicates the Waymo training set is very well balanced.} 
  \label{fig:bdd100k-samples-w-likelihood}

\end{figure*}

\newpage
\section{BDD100K training dataset}\label{sec:bdd100k-supplementary}

\begin{figure*}[!h]  
  \centering
  \subfloat[$\mathcal{L}_{s^{(i)}}=0.0049$]
  {\includegraphics[width=0.9\linewidth]{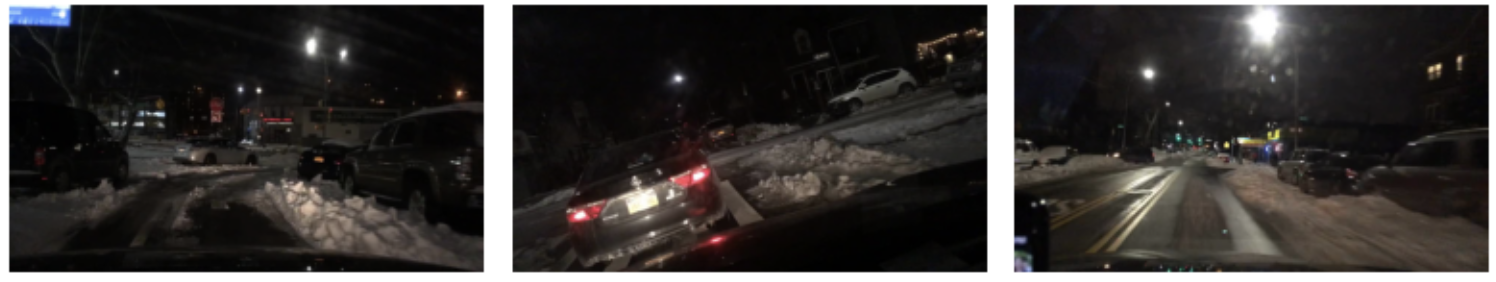}}

  \centering
  \subfloat[$\mathcal{L}_{s^{(i)}}=0.0083$]
  {\includegraphics[width=0.9\linewidth]{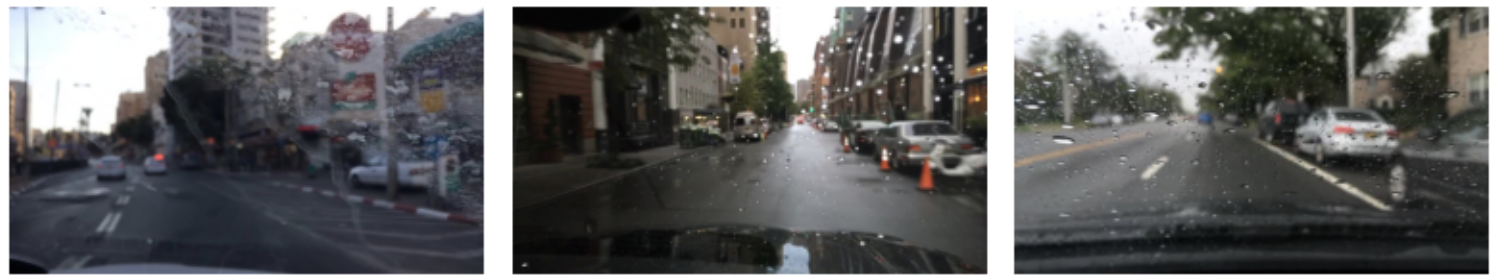}}

  \centering
  \subfloat[$\mathcal{L}_{s^{(i)}}=0.0086$]
  {\includegraphics[width=0.9\linewidth]{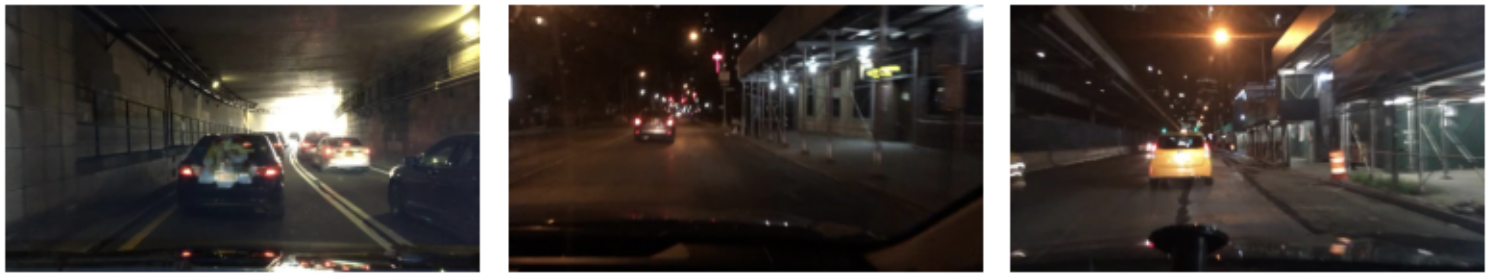}}
  
  \centering
  \subfloat[$\mathcal{L}_{s^{(i)}}=0.7268$]
  {\includegraphics[width=0.9\linewidth]{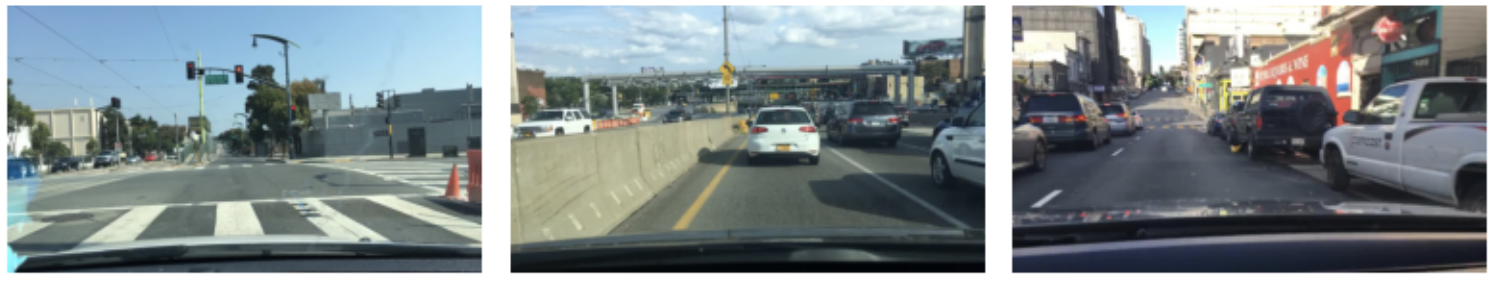}}
  
  \centering
  \subfloat[$\mathcal{L}_{s^{(i)}}=1.0$]
  {\includegraphics[width=0.9\linewidth]{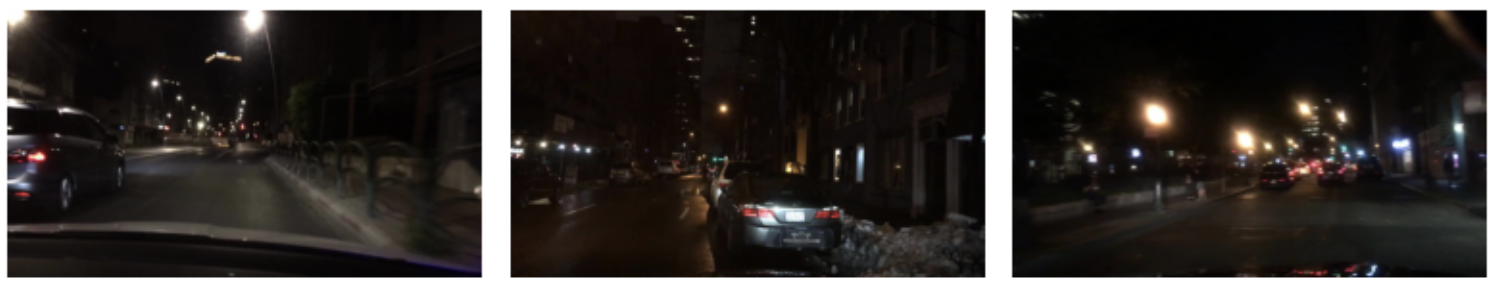}}

  \caption{Samples from various clusters in BDD100K \textit{training} dataset, and their likelihoods relative to the largest cluster within the dataset. Samples shown in the last row come from a cluster of size $23385$, and is the largest among all clusters. The smallest cluster shown in the top row contains only $116$ samples.}
  \label{fig:bdd100k-samples-w-likelihood}
\end{figure*}

\end{document}